\documentclass[accepted]{uai2025} % after acceptance, for a revised version; 

\usepackage[american]{babel}

\usepackage{natbib} %
\usepackage{mathtools} %
\usepackage{booktabs} %
\usepackage{tikz} %

\usepackage[utf8]{inputenc} %
\usepackage[T1]{fontenc}    %
\usepackage{hyperref}       %
\usepackage{url}            %
\usepackage{amsfonts}       %
\usepackage{nicefrac}       %
\usepackage{microtype}      %
\usepackage{xcolor}         %

\usepackage{amsmath,amsfonts,bm,amsthm}
\usepackage{amssymb}
\usepackage{caption}
\usepackage{subcaption}
\usepackage{pifont}

\title{Hiding and Recovering Knowledge in Text-to-Image \\ Diffusion Models via Learnable Prompts}

\author[1]{\href{mailto:tuananh.bui@monash.edu}{Anh Bui}}
\author[2]{Khanh Doan}
\author[1]{Trung Le}
\author[3]{Paul Montague}
\author[3]{Tamas Abraham}
\author[1]{Dinh Phung}
\affil[1]{%
    Monash University\\
}
\affil[2]{%
    VinAI Research\\
}
\affil[3]{%
    Defence Science and Technology Group\\
  }
  
\begin{document}
\maketitle

\begin{abstract}
Diffusion models have demonstrated remarkable capability in generating high-quality visual content from textual descriptions. However, since these models are trained on large-scale internet data, they inevitably learn undesirable concepts, such as sensitive content, copyrighted material, and harmful or unethical elements. While previous works focus on permanently removing such concepts, this approach is often impractical, as it can degrade model performance and lead to irreversible loss of information.
In this work, we introduce a novel concept-hiding approach that makes unwanted concepts inaccessible to public users while allowing controlled recovery when needed. Instead of erasing knowledge from the model entirely, we incorporate a learnable prompt into the cross-attention module, acting as a secure memory that suppresses the generation of hidden concepts unless a secret key is provided. This enables flexible access control—ensuring that undesirable content cannot be easily generated while preserving the option to reinstate it under restricted conditions.
Our method introduces a new paradigm where concept suppression and controlled recovery coexist, which was not feasible in prior works. We validate its effectiveness on the Stable Diffusion model, demonstrating that hiding concepts mitigates the risks of permanent removal while maintaining the model’s overall capability.
% Our code is anonymously available at \url{https://anonymous.4open.science/r/Erasing-KPOP/}.
Our code is available at \url{https://github.com/tuananhbui89/Erasing-KPOP}.
\end{abstract}

\section{Introduction}

Recent advances in text-to-image generative models \citep{rombach2022high, ramesh2021zero, ramesh2022hierarchical} have garnered significant attention due to their exceptional image quality and creative potential. These models are trained on massive internet-scale datasets, enabling them to generate diverse visual content. However, these datasets often contain undesirable concepts—such as sensitive, harmful, or copyrighted material—that may be inadvertently learned by the models. As a result, users can potentially exploit these models to generate content that spreads misinformation, hate speech, or other unethical material \citep{rando2022red, qu2023unsafe, westerlund2019emergence}. To ensure the safety and responsible deployment of these models, it is crucial to prevent the generation of such undesirable content.

To address this challenge, prior works have explored various strategies, including dataset filtering \citep{SD20}, post-generation filtering \citep{rando2022red}, and inference-time guidance \citep{schramowski2023safe}. However, each of these approaches has significant limitations. 
Specifically, dataset filtering requires extensive human effort to curate training data and has been shown to degrade model performance on general tasks \citep{SD20}, 
while post-generation filtering (e.g., NSFW classifiers \citep{SD20}) struggles with false positives and can be easily bypassed by users.
Inference-time guidance provides a lightweight solution but can be overridden by adversarial attacks \citep{gandikota2023erasing}. 

A more effective approach is to modify the model itself to remove unwanted concepts. Existing methods \citep{gandikota2023erasing, kumari2023ablating, orgad2023editing, zhang2023forget, bui2024erasing, bui2025fantastic} attempt to erase knowledge of undesirable concepts through fine-tuning, optimizing specific loss functions to suppress them. 
However, this permanent removal of concepts has inherent drawbacks. 
Firstly, cince model parameters are highly shared across concepts, erasing one concept can unintentionally degrade related capabilities.
Secondly, once a concept is removed, recovering it requires retraining the model, which may not be feasible in real-world scenarios. 

In this work, we introduce a fundamentally different approach: rather than permanently removing, we propose to \textbf{hide undesirable concepts} within the model. 
Our method prevents public users from generating unwanted content while allowing controlled recovery using a secret key. This paradigm offers key advantages over previous approaches, 
such as \textbf{flexibility and controlled access} where hidden concepts can be reinstated as needed, enabling adaptive moderation.

Inspired by prompt-based tuning techniques \citep{Li2021,Lester2021, Pfeiffer2020}, we introduce a learnable prompt to hide undesirable concepts. 
This prompt serves as an additional memory module that captures knowledge of undesirable concepts, reducing their dependence on core model parameters. 
High-level speaking, our method consists of two stages: \textbf{knowledge recovery/transfer} - to transfer knowledge of undesirable concepts to the prompt, 
and \textbf{knowledge hiding/removal} - to hide the undesirable concepts from the model.

Through extensive experiments, we demonstrate that our approach successfully prevents the generation of undesirable content while maintaining the model’s general capabilities. We evaluate this across three scenarios:
\ding{192} Hiding object-oriented concepts,
\ding{193} Mitigating unethical content, and
\ding{194} Hiding artistic style concepts.
Our results highlight the practical impact of concept hiding, paving the way for more adaptive and secure content moderation in text-to-image models.

\section{Related Work}

In this section, we provide an extensive overview of existing literature concerning concept erasure and related techniques to our work.

\paragraph*{Concept Erasing Techniques:}

We categorize concept erasing techniques into four main classes: (1) Pre-processing, (2) Post-processing, (3) Anti Concept Mimicry, and (4) Model Editing.

Pre-processing methods represent a straightforward approach to eliminating undesired concepts from input images. This involves employing pre-trained detectors to identify images containing objectionable content and subsequently excluding them from the training set. However, the drawback lies in the necessity of retraining the model from scratch, which proves computationally expensive and impractical for evolving erasure requests. A notable instance of complete retraining is evident in Stable Diffusion v2.0 \citep{SD20}, but this approach was reported to leave the model inadequately sanitized \citep{gandikota2023erasing}.

Post-processing methods encompass the utilization of Not-Safe-For-Work (NSFW) detectors to identify potentially inappropriate content in generated images. Images flagged by the NSFW detector are then either blurred or blacked out before being presented to users. This method, employed by organizations such as OpenAI (developer of Dall-E), StabilityAI (developer of Stable Diffusion), and Midjourney Inc (developer of Midjourney), is considered highly effective. However, the open-source nature of the Stable Diffusion model exposes it to potential evasion by modifying the NSFW detector in the source code. Closed-source models, like Dall-E, are not immune either, as demonstrated in \citep{yang2024sneakyprompt}, where a technique similar to Boundary Attack \citep{brendel2017decision} was used to uncover adversarial prompts that could bypass the filtering mechanism.

Concept Mimicry serves as a personalization technique, generating images aligned with a user's preferences based on their input. Noteworthy methods include Textual Inversion \citep{gal2022image} and Dreambooth \citep{ruiz2023dreambooth}, which have proven effective with minimal user input. In contrast, Anti Concept Mimicry is employed to safeguard personal or artistic styles from being copied through Concept Mimicry. Achieved by introducing imperceptible adversarial noise to input images, this technique can deceive Concept Mimicry methods under specific conditions. Recent contributions such as Anti-Dreambooth \citep{van2023anti} have explored and demonstrated the effectiveness of this approach.

To date, the most successful strategy for sanitizing open-source models, such as Stable Diffusion, involves cleaning the generator (e.g., U-Net) in the diffusion model post-training on raw, unfiltered data and before public release. This approach, as partially demonstrated in \citet{gandikota2023erasing}, underscores the importance of addressing potential biases and undesired content in models before their deployment.

\paragraph*{Existing erasing methods:}

Latent Diffusion models (LDMs) are combined techniques to control generated images by input text. The encoder and decoder of a variational autoencoder (VAE) model are used to bring input from pixel space into latent space and from U-Net model output in reverse. Meanwhile, text is embedded by a pre-trained CLIP model. Cross-Attention is the way to align context from text embedding into image information flow. From that, several existing works show that fine-tuning the Cross-Attention layer only (linear projection layers of key and value) or Text Encoder only or both of them are sufficient ways to customize a pre-trained LDM.

Existing erasing methods \citep{gandikota2023erasing, orgad2023editing, zhang2023forget, kumari2023ablating} aim to erase undesirable concepts by fine-tuning foundation models with appropriate losses to unlearn and erase these undesirable concepts. Specifically, TIME \citep{orgad2023editing}, UCE \citep{zhang2023forget}, Concept Ablation \citep{kumari2023ablating}, and SDD try to project meaning of harmful context into another benign one, while ESD \citep{gandikota2023erasing} uses the principle of classifier-free-guidance to remove the distribution of the bad concept from the LDM.

\paragraph*{Prompting for transfer learning:} The overarching concept behind prompting involves applying a function to alter the input text, providing the language model with supplementary task-related information. However, devising an effective prompting function presents challenges and necessitates heuristic approaches. Recent studies, such as prompt tuning \citep{Lester2021} and prefix tuning \citep{Li2021}, attempt to tackle this challenge by employing trainable prompts within a continuous space, resulting in impressive performance in transfer learning tasks. Prompts encapsulate task-specific knowledge with significantly fewer additional parameters compared to competing methods like Adapter \citep{Pfeiffer2020, Wang2021a} and LoRA \citep{Hu2022}.

\section{Proposed Method}

\subsection{Motivation}
\label{sec:motivation_hiding}

Unlearning unwanted concepts, though initially explored, has recently gained significant attention, with several notable works \citep{gandikota2023erasing, gandikota2024unified, zhang2023forget, bui2024erasing, bui2025fantastic} demonstrating that it is possible to remove specific concepts from foundation models before their deployment.

However, in this work, we pursue a different objective compared to existing approaches. Instead of permanently removing unwanted concepts, we aim to hide them—preventing public users from generating undesirable content while allowing model developers to later recover these concepts using a secret key known only to them.

This new hiding paradigm is particularly important in real-world applications for several reasons:

- \textbf{Flexibility \& Controlled Access:} Unlike methods that permanently remove concepts, hiding allows model developers to reinstate specific concepts if regulations evolve, new use cases arise, or controlled access is required. This approach also enables the development of novel business models where different levels of access can be granted.

- \textbf{Enhancing Security Against Backdoor Attacks:} Our method can be viewed as a deliberately introduced, controlled backdoor for beneficial purposes. By studying the feasibility of embedding and later recovering hidden concepts, developers can gain insights into potential security vulnerabilities in text-to-image (T2I) models and proactively implement safeguards.

\subsection{Knowledge Hiding and Recovery with Prompt - KPOP}

\paragraph{High-level Overview}

Given a pre-trained foundation model $\theta$ and a set of undesirable concepts $\mathbf{E}$, we propose a fine-tuning process to obtain a sanitized model $\theta^{'}$ that satisfies the two key objectives: \textbf{hiding} - the undesirable concepts are concealed from $\theta^{'}$ preventing public users from generating them, and \textbf{recovery} - the hidden concepts can be reinstated when a secret key is provided.

Mathematically, let $\epsilon_{\theta}(z_t, c, t)$ denote the output of the \textit{pre-trained foundation} U-Net model parameterized by $\theta$ at step $t$ given an input description $c$ and the latent vector from the previous step $z_t$. 
Similarly, let $\epsilon_{\theta^{'}}(z_t, c, t)$ denote the output of the \textit{sanitized} model, parameterized by parameters $\theta^{'}$.  
For simplicity, we employ the notations $\epsilon_{\theta}(c)$ and $\epsilon_{\theta^{'}}(c)$. 
The \textbf{hiding} goal can be formulated as $\epsilon_{\theta^{'}}(c) \neq \epsilon_{\theta}(c) \; \forall c \in \mathbf{E}$, 
while the \textbf{recovery} goal can be represented as $\epsilon_{\theta^{'}}(c, \mathbf{p}_{c_e}) \approx \epsilon_{\theta}(c) \; \forall c \in \mathbf{E}$, 
where $\mathbf{p}_{c_e}$ is a learnable prompt acting as the \textbf{secret key} to recover the concept $c_e \in \mathbf{E}$.

To achieve the two objectives, we propose a fine-tuning process that alternates between two stages: \textbf{knowledge recovery/transfer} and \textbf{knowledge hiding/removal} integrated within the following optimization problem:

\begin{equation}
    \label{eq:main_objective}
    \underset{\theta^{'}}{\min}\;\mathbb{E}_{c_{e}\in\mathbf{E}}\left[\underbrace{\left\Vert \epsilon_{\theta^{'}}(c_{e})-\epsilon_{\theta}(c_{t})\right\Vert _{2}^{2}}_{L1}+\lambda\underbrace{\left\Vert \epsilon_{\theta^{'}}(c_e,\mathbf{p}_{c_e})-\epsilon_{\theta}(c_{e})\right\Vert _{2}^{2}}_{L2}\right]
\end{equation}

\paragraph{Knowledge Hiding/Removal}

The goal of this stage is to hide the undesirable concepts $c_e \in \mathbf{E}$ from the model $\theta^{'}$. 
This can be done by optimizing the OP \ref{eq:main_objective}, 
where minimizing the $L_1$ loss enforces the model's output for undesirable concepts closely resembles that of a target concept $c_t$ which is usually a neutral concept like `a photo' or empty description ` ' as in previous works \citep{gandikota2023erasing, bui2024erasing}.
However, minimizing only the $L_1$ loss often leads to severe degradation in unrelated concepts. 
To address this, we simultaneously minimize the $L_2$ loss which enables the recovery of hidden concepts using an additional prompt $\mathbf{p}_{c_e}$ to guide the recovery process.
This guided recovery process reduces the model’s reliance on standard textual inputs, effectively transferring the knowledge of hidden concepts to the prompt $\mathbf{p}_{c_e}$.

\paragraph{Knowledge Recovery/Transfer}

While the objective of the \textbf{Knowledge Hiding} stage has been previously proposed in \citep{gandikota2023erasing, bui2024erasing}, 
our main contribution lies in the \textbf{Knowledge Recovery} stage which really makes the hidden knowledge recovery possible rather than being removed as in previous works. 
Since the fine-tuned model $\theta^{'}$ has been altered from the original model $\theta$ due to the knowledge hiding process, 
the goal of this stage is to find the prompt $\mathbf{p}_{c_e}$ that can recover the undesirable concept $c_e$ that has been removed from the model $\theta^{'}$.
This is formulated as the following optimization problem \ref{eq:hidden} \citep{gal2022image}.

\begin{equation}
    \label{eq:hidden}
    \mathbf{p}_{c_e} = \underset{\mathbf{p}:\Vert \mathbf{p} - c_e \Vert_{2}\leq\rho}{\text{argmin}} \;  \left\| \epsilon_{\theta^{'}}(c_e, \mathbf{p}) - \epsilon_{\theta}(c_e) \right\|_2^2
\end{equation}

where $\mathbf{p}_{c_e}$ is the learnable prompt, associated with the undesirable concept $c_e$, and $\rho$ is the constraint on the prompt to make it not too far from the undesirable concept $c_e$. 
The solution $\mathbf{p}_{c_e}$ can be efficiently leanred via gradient descent. 

\paragraph{Implication on Backdooring T2I models}. 

Since the sanitized model $\theta^{'}$ satisfies $\epsilon_{\theta^{'}}(c_e,\mathbf{p}_{c_e}) \approx \epsilon_{\theta}(c_e)$ while $\epsilon_{\theta^{'}}(c_e) \neq \epsilon_{\theta}(c_e)$,
the prompt $\mathbf{p}_{c_e}$ effectively acts as a \textbf{backdoor trigger} for the sanitized model $\theta^{'}$. 
This means that, even though model $\theta^{'}$ has been trained to hide the concept $c_e$, applying the secret key $\mathbf{p}_{c_e}$ can restore its generation capability of the undesirable concept $c_e$.

As demonstrated in Section \ref{sec:object_related_concepts}, the quality of the recovered generated images with the key $\mathbf{p}_{c_e}$ is comparable to those produced by the original model $\theta$. 
Thus, the secret key $\mathbf{p}_{c_e}$ serves as a \textit{hidden access mechanism}, allowing the model developers to flexibly control the model's generation capability or analyze the model's vulnerability to future backdoor attacks.
We provide a detailed description of our method in Section \ref{sec:further_details}.

\section{Experiments}
\label{sec:experiments}

\begin{table*}[h]
    \centering
    \caption{Erasing object-related concepts. Ours$^{\star}$ denote the results with the setting with the knowledge of to-be-preserved concepts.}
    \resizebox{0.80\textwidth}{!}{%
    \begin{tabular}{cccccccccc}
        Method & ESR-1$\uparrow$ & ESR-5$\uparrow$ & PSR-1$\uparrow$ & PSR-5$\uparrow$ & RSR-1$\uparrow$ & RSR-5$\uparrow$\tabularnewline
        \hline 
        SD & $22.0\pm11.6$ & $2.4\pm1.4$ & $78.0\pm11.6$ & $97.6\pm1.4$ & N/A & N/A\tabularnewline
        ESD & $95.5\pm0.8$ & $88.9\pm1.0$ & $41.2\pm12.9$ & $56.1\pm12.4$ & N/A & N/A\tabularnewline
        CA & $98.4\pm0.3$ & $96.8\pm6.1$ & $44.2\pm9.7$ & $66.5\pm6.1$ & N/A & N/A\tabularnewline
        UCE & $100\pm0.0$ & $100\pm0.0$ & $23.4\pm3.6$ & $49.5\pm8.0$ & N/A & N/A\tabularnewline
        UCE$^{\star}$ & $100\pm0.0$ & $100\pm0.0$ & $62.1\pm34.6$ & $96.0\pm2.9$ & N/A & N/A\tabularnewline
        \hline
        Ours & $99.5\pm0.3$ & $98.0\pm1.9$ & $26.6\pm5.7$ & $47.8\pm5.0$ & $72.0\pm11.2$ & $97.2\pm2.4$ \tabularnewline
        Ours$^{\star}$ & $99.2\pm0.5$ & $97.3\pm1.9$ & $75.3\pm12.0$ & $98.0\pm0.5$ & $71.5\pm9.7$ & $95.3\pm3.6$ \tabularnewline
        \hline 
    \end{tabular}
    }
    \label{tab:object_erasing}
\end{table*}

\subsection{General Settings}
\label{sec:general_settings}

Standard evaluation protocol for erasing/unlearning concepts in literature is to evaluate the performance of the sanitized model on erasing tasks and preserving tasks.
In this paper, because of the new capability of our method, we propose a new additional metric to evaluate the recovering performance of the sanitized model with the secret key.

More specifically, we conduct three sets of experiments, i.e., erasing object-related concepts, erasing artistic style concepts, and erasing unethical content using Stable Diffusion (SD) version 1.4 as the foundation model. 
We employ the same setting across all methods, i.e., fine-tuning the model for 1000 steps with a batch size of 1, with the Adam optimizer with a learning rate of $1\mathrm{e}{-5}$.
We benchmark our method against four baseline approaches, namely, the original pre-trained SD model, ESD \citep{gandikota2023erasing}, UCE \citep{gandikota2024unified}, and Concept Ablation (CA) \citep{kumari2023ablating}.

\subsection{Erasing Object-Related Concepts}
\label{sec:object_related_concepts}

In this experiment, we assess our method's capability to remove/hide object-related concepts from the foundation model, such as erasing entire object classes like `Cassette Player'.
We follow the setting in \citet{gandikota2023erasing,bui2024erasing} and use the Imagenette \footnote{https://github.com/fastai/imagenette}, a subset of the ImageNet dataset \citep{imagenet}, which comprises 10 easily recognizable classes, including `Cassette Player', `Chain Saw', `Church', `Gas Pump', `Tench', `Garbage Truck', `English Springer', `Golf Ball', `Parachute', and `French Horn'.
We erase a set of 5 classes simultaneously, where we generate 500 images for each class and use the pre-trained ResNet-50 \citep{resnet} to detect the presence of an object in these images.
We employ three metrics: \textbf{Erasing Success Rate (ESR-k)}: The percentage of generated images with `to-be-erased' classes where the object is not detected in the top-k predictions. 
\textbf{Preserving Success Rate (PSR-k)}: That with `to-be-preserved' classes where the object is detected in the top-k predictions.
\textbf{Recovering Success Rate (RSR-k)}: That with `to-be-erased' classes when the secret key is provided.

\paragraph*{Removing and Preserving Performance.}

We select four distinct sets of five classes from the Imagenette dataset for erasure and present the results in Table \ref{tab:object_erasing}.
First, we note that the average PSR-1 and PSR-5 scores across the four settings of the original SD model stand at 78.0\% and 97.6\%, respectively.
This means that 78.0\% of the generated images contain the object-related concepts, which are detected in the top-1 prediction, 
and this number increases to 97.6\% when considering the top-5 predictions. 
These scores highlight the original SD model's ability to generate images with the expected object-related concepts. 

Regarding erasing performance, all baselines achieve very high ESR-1 and ESR-5 scores, with the lowest being 95.5\% and 88.9\%, respectively.
This demonstrates the effectiveness of these methods in erasing object-related concepts, as only a small proportion of the generated images contained the concepts upon detection. 
Notably, the UCE method achieves 100\% ESR-1 and ESR-5, the highest among the baselines.
Our method achieves 99.2\% ESR-1 and 97.3\% ESR-5, which is much higher than the two baselines ESD and CA, and only slightly lower than the UCE method, which is designed specifically for erasing object-related concepts.

However, despite the high erasing performance, the baselines suffer from a significant drop in preserving performance, 
with the lowest PSR-1 and PSR-5 scores being 41.2\% and 56.1\%, respectively. 
This suggests that the preservation task is more challenging, and the baselines are ineffective in retaining other concepts. 
In contrast, our method achieves 75.3\% PSR-1 and 98.0\% PSR-5, which is a significant improvement compared to the best baseline, UCE, with 62.1\% PSR-1 and 96.0\% PSR-5. 
Compared to the same setting without the knowledge of to-be-preserved concepts (denoted as UCE$^{\star}$ and Ours$^{\star}$), 
our method still achieves competitive results, with 3.2\% higher PSR-1 but 1.7\% lower PSR-5 than UCE$^{\star}$. 
This result underscores the effectiveness of our method in simultaneously erasing object-related concepts while preserving other unrelated concepts.

\begin{figure}
    \centering
    \includegraphics[width=\columnwidth]{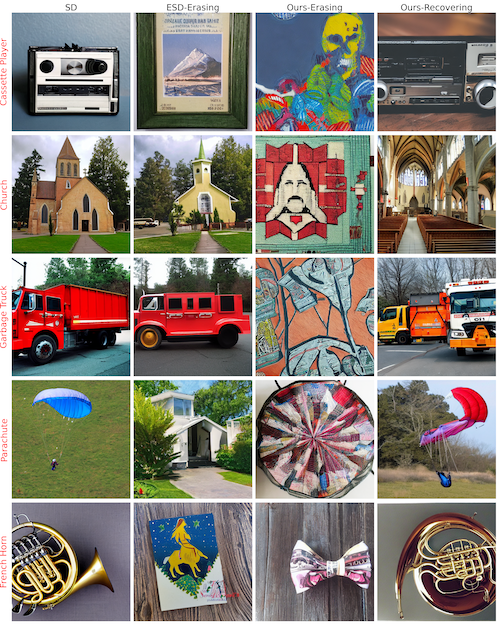}
    \caption{Qualitative results. (1st column) Original SD model. (2nd column) Sanitized by ESD. (3rd column) Sanitized by our method. (4th column) Recovered by our method using the secret key.
    Several failure cases in ESD demonstrate incomplete erasure, whereas our approach effectively removes the target concepts while enabling precise recovery with the secret key.}
    \label{fig:recover}
\end{figure}

\paragraph*{Recovering Performance.}

In this section, we highlight a unique property of our method: the ability to \textbf{recover hidden concepts} from the sanitized model using a secret key.

As shown in Table \ref{tab:object_erasing}, our approach achieves \textbf{71.5\% RSR-1 and 95.3\% RSR-5}, closely matching the original SD model's \textbf{78.0\% and 97.6\%} success rates for top-1 and top-5 predictions, respectively.
This demonstrates that our method effectively preserves the hidden knowledge while restricting unauthorized access.

For qualitative evaluation, Figure \ref{fig:recover} presents a visual comparison of generated images across different models. 
The first column shows the original images produced by the SD model, while the second and third columns depict sanitized outputs generated by ESD and our method, respectively. 
The fourth column illustrates the recovered images using our secret key mechanism.

Our method successfully removes the undesirable concepts from the public model while still allowing retrieval when needed. 
Notably, ESD fails to completely erase certain concepts, leading to inconsistencies in its sanitization process. 
In contrast, our approach not only ensures effective concept removal but also enables seamless recovery via the secret key, as evidenced by the high-quality regenerated images in Figure \ref{fig:recover}.

These results confirm that the knowledge of hidden concepts remains intact within the sanitized model but is inaccessible to the public through standard textual prompts, 
ensuring both security and controlled access. This capability is crucial for scenarios where selective access to sensitive information is required while maintaining the integrity of the generative model.

\paragraph*{Visualizing Attribution Maps.}

To gain deeper insights into the behavior of our method, we leverage DAAM \citep{tang2022daam} to visualize the attentive attribution maps that depict the interaction between visual and textual concepts in the generated images. 
DAAM is an emerging technique that interprets how an input word influences parts of the generated image by analyzing the attention maps in the cross-attention module of the Stable Diffusion model. 

\begin{figure*}
    \centering
    \includegraphics[width=\textwidth]{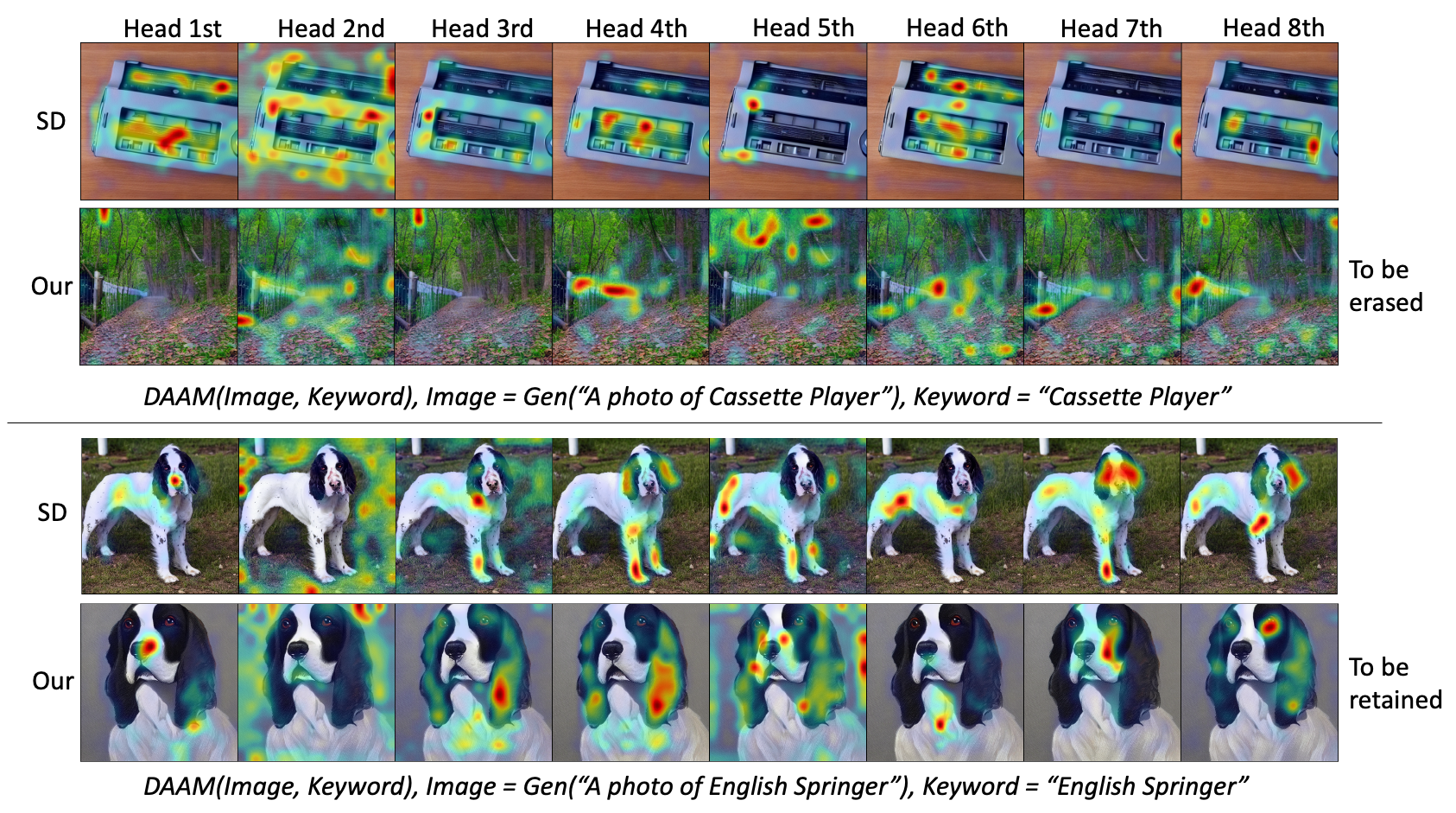}
    \caption{Attentive attribution maps between the visual and textual concepts in the original SD model and our method.}
    \label{fig:daam_attention}
\end{figure*}

We first use DAAM to analyze the original SD model's behavior in generating images with  `Cassette Player' and `English Springer' as input prompts as shown in Figure \ref{fig:daam_attention}. 
Each test case comprises eight sub-figures, each of which corresponds to a head in the multihead cross-attention module.
As depicted in Figure \ref{fig:daam_attention}, most heatmaps concentrate on the cassette player and the dog's body, aligning well with the respective textual prompts. 
Interestingly, the second head does not focus on the cassette player or the dog's body but instead on the surrounding background. 

We then utilize DAAM to visualize the attribution maps of generated images using the same prompts with our method. 
We find that on the concept to be retained (i.e., `English Springer'), the heatmaps also focus on the dog's body except for the second head, 
mirroring the behavior observed in the original SD model. 
For the concept to be erased (i.e., `Cassette Player'), the heatmaps exhibit a more dispersed pattern, indicating a lack of specific concentration on any distinct region. 
This observation suggests that the model, under the erasure effect of our method, diverts attention away from the cassette player concept, 
providing valuable insights into the underlying mechanism of our method.

\subsection{Mitigating Unethical Content}
\label{sec:unethical_content}

\paragraph*{Setting.}

In this section, we evaluate the effectiveness of our method in erasing/hiding NSFW content, including sexual, violent, and racist content. 
Specifically, we follow the setting in \citet{gandikota2023erasing, gandikota2024unified,bui2024erasing}, we fine-tune the non-cross-attention modules, and using the keyword `nudity' as the input prompt to identify the NSFW content.
We use the I2P dataset \citep{schramowski2023safe} to generate a set of 4703 images containing attributes of sexual, violent, and racist content. 
We utilize the NudeNet detector \citep{nudenet2019}, which accurately detects various exposed body parts, to identify the presence of nudity in these images.
The NudeNet detector provides multi-label predictions with associated confidence scores, allowing us to adjust the threshold and control the trade-off between the number of detected body parts and the confidence of the detection—higher thresholds result in fewer detected body parts.

\paragraph*{Experimental Results.}

\begin{figure}
    \centering
    \begin{subfigure}{0.6\columnwidth}
        \centering
        \includegraphics[width=\columnwidth]{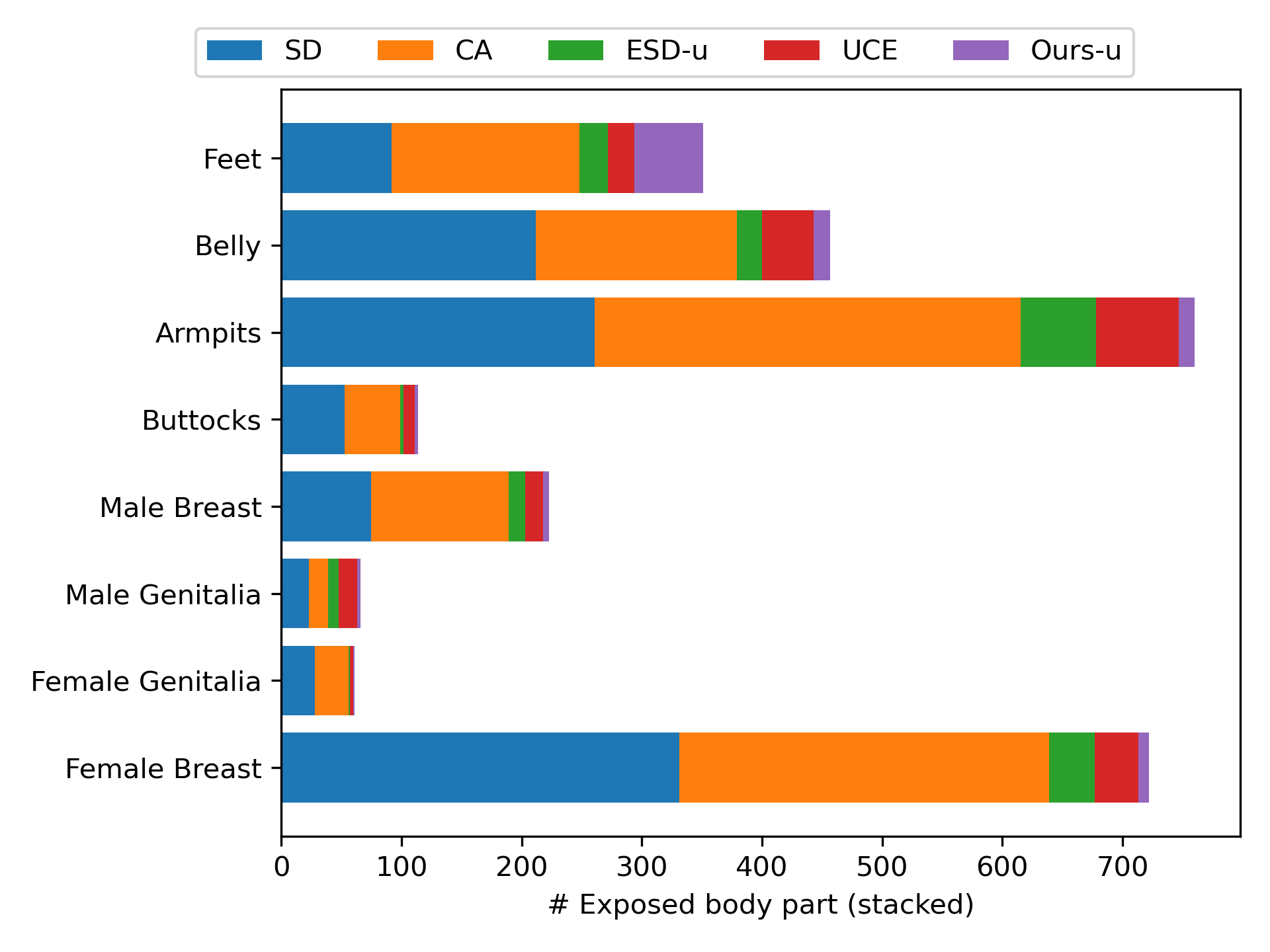}
        \caption{}
        \label{fig:exposed_body_parts_stacked}
    \end{subfigure}

    \begin{subfigure}{0.6\columnwidth}
        \centering
        \includegraphics[width=\columnwidth]{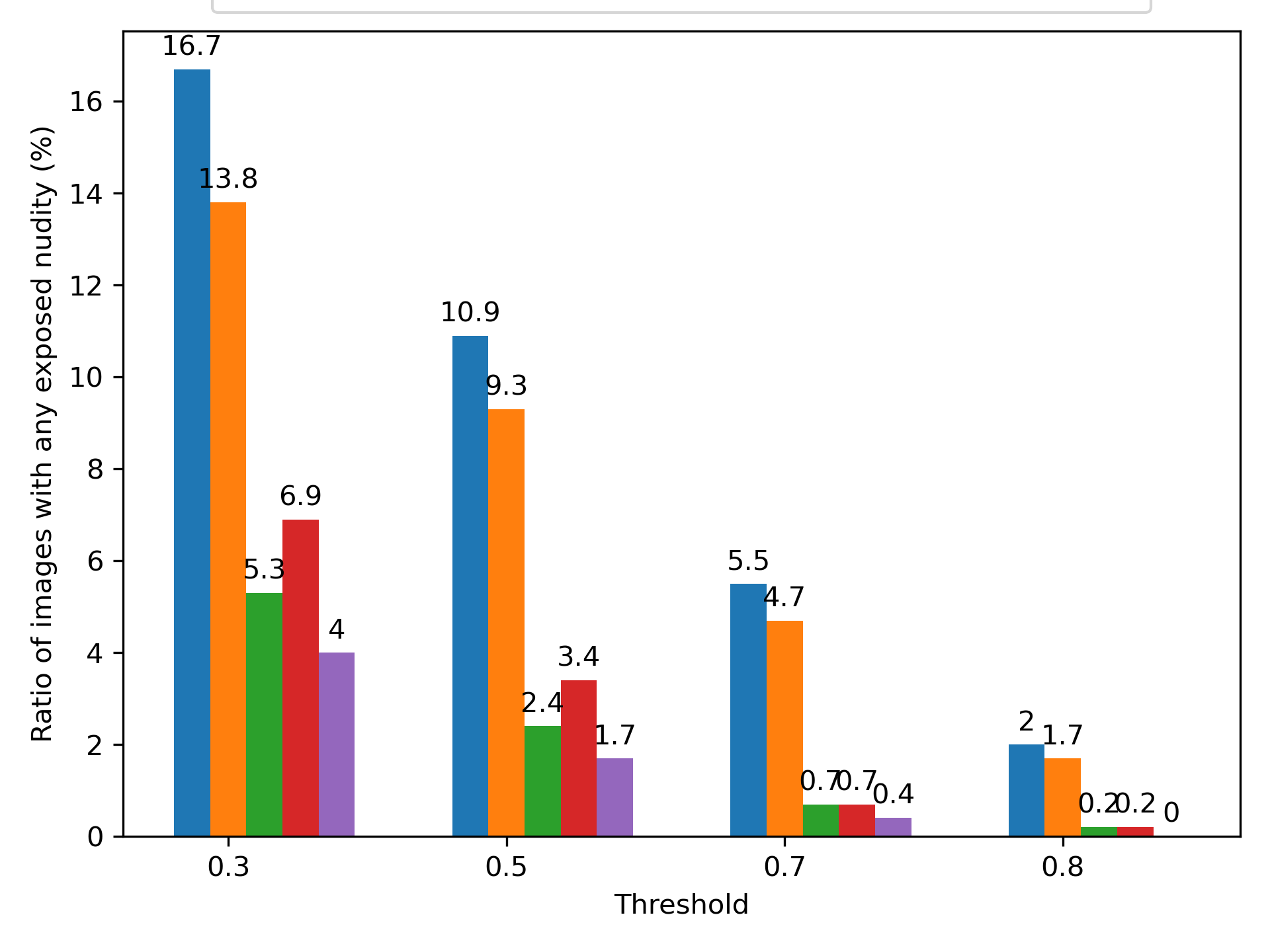}
        \caption{}
        \label{fig:exposed_nudity}
    \end{subfigure}
    \caption{Comparison of the erasing performance on the I2P dataset. \ref{fig:exposed_body_parts_stacked}: Number of exposed body parts counted in all generated images with threshold 0.5. 
    \ref{fig:exposed_nudity}: Ratio of images with any exposed body parts detected by the detector \citep{nudenet2019}.}
    \label{fig:combine_exposed_nudity}
\end{figure}

Figure \ref{fig:exposed_body_parts_stacked} shows the ratio of images with any exposed body parts detected by the detector \citep{nudenet2019} across the total 4703 generated images (denoted by \textbf{NER}) across thresholds ranging from 0.3 to 0.8. 
Notably, our method consistently outperforms the baselines under all thresholds, demonstrating its effectiveness in erasing NSFW content. 
Specifically, with the threshold set at 0.3, the NER score for the original SD model stands at 16.7\%, indicating that 16.7\% of the generated images contain signs of nudity concept. 
The two baselines, ESD and UCE, achieve 5.32\% and 6.87\% NER with the same threshold, respectively, demonstrating their effectiveness in erasing nudity concepts. 
Our method achieves a NER score of 3.95\%, the lowest among the baselines, indicating the highest erasing performance. 
This result remains consistent across different thresholds, emphasizing the robustness of our method in erasing NSFW content.

Additionally, to measure the preserving performance, we generate images with COCO 30K prompts and measure the FID score compared to COCO 30K validation images. 
Our method achieves an FID score of 16.73, slightly lower than that of UCE, which is the highest score at 15.98, indicating that our method can simultaneously erase a concept while preserving other concepts effectively.

Detailed statistics of different exposed body parts in the generated images are provided in Figure \ref{fig:exposed_nudity}.
It can be seen that in the original SD model, among all the body parts, the female breast is the most detected body part in the generated images, accounting for more than 320 images out of the total 4703 images. 
Both baselines, ESD and UCE, as well as our method, achieve a significant reduction in the number of detected body parts, with our method achieving the lowest number among the baselines.
Our method also achieves the lowest number of detected body parts for the most sensitive body parts, only surpassing the baseline for less sensitive body parts, such as feet.

\begin{table}[h]
    \centering
    \caption{Evaluation on the nudity erasure setting.}
    \resizebox{1.0\columnwidth}{!}{%
    \begin{tabular}{cccccc}
     & NER-0.3$\downarrow$ & NER-0.5$\downarrow$ & NER-0.7$\downarrow$ & NER-0.8$\downarrow$ & FID$\downarrow$\tabularnewline
    \hline 
    CA & 13.84 & 9.27 & 4.74 & 1.68 & 20.76\tabularnewline
    UCE & 6.87 & 3.42 & 0.68 & 0.21 & 15.98\tabularnewline
    ESD & 5.32 & 2.36 & 0.74 & 0.23 & 17.14\tabularnewline
    \hline 
    Ours & 3.95 & 1.70 & 0.40 & 0.0 & 16.73\tabularnewline
    \hline 
    \end{tabular}
    }
    \label{tab:nudity_coco}
\end{table}

\subsection{Erasing Artistic Style Concepts}
\label{sec:artistic_style_concepts}

\paragraph*{Setting.}
In this experiment, we investigate the ability of our method to erase/hide artistic style concepts. 
We select several famous artists with easily recognizable styles who have been known to be mimicked by the text-to-image generative models, 
including "Kelly Mckernan", "Thomas Kinkade", "Tyler Edlin" and "Kilian Eng" as in \citet{gandikota2023erasing}.
We compare our method with recent work including ESD \citep{gandikota2023erasing}, UCE \citep{gandikota2024unified}, and CA \citep{kumari2023ablating} 
which have demonstrated effectiveness in similar settings.

\paragraph*{Experimental Results.}

For fine-tuning the model, we use only the names of the artists as inputs. 
For evaluation, we use a list of long textual prompts that are designed exclusively for each artist, combined with 5 seeds per prompt to generate 200 images for each artist across all methods.  
We measure the CLIP alignment score \footnote{https://lightning.ai/docs/torchmetrics/stable/multimodal/\\clip\_score.html} between the visual features of the generated images and their corresponding textual embeddings. 
Compared to the setting \citep{gandikota2023erasing} which used a list of generic prompts, our setting with longer, specific prompts can leverage the CLIP score as a more meaningful measurement to evaluate the erasing and preserving performance. 
We also use LPIPS \citep{zhang2018unreasonable} to measure the distortion in generated images by the original SD model and editing methods, where a low LPIPS score indicates less distortion between two sets of images. 

\begin{figure}
    \centering
    \includegraphics[width=\columnwidth]{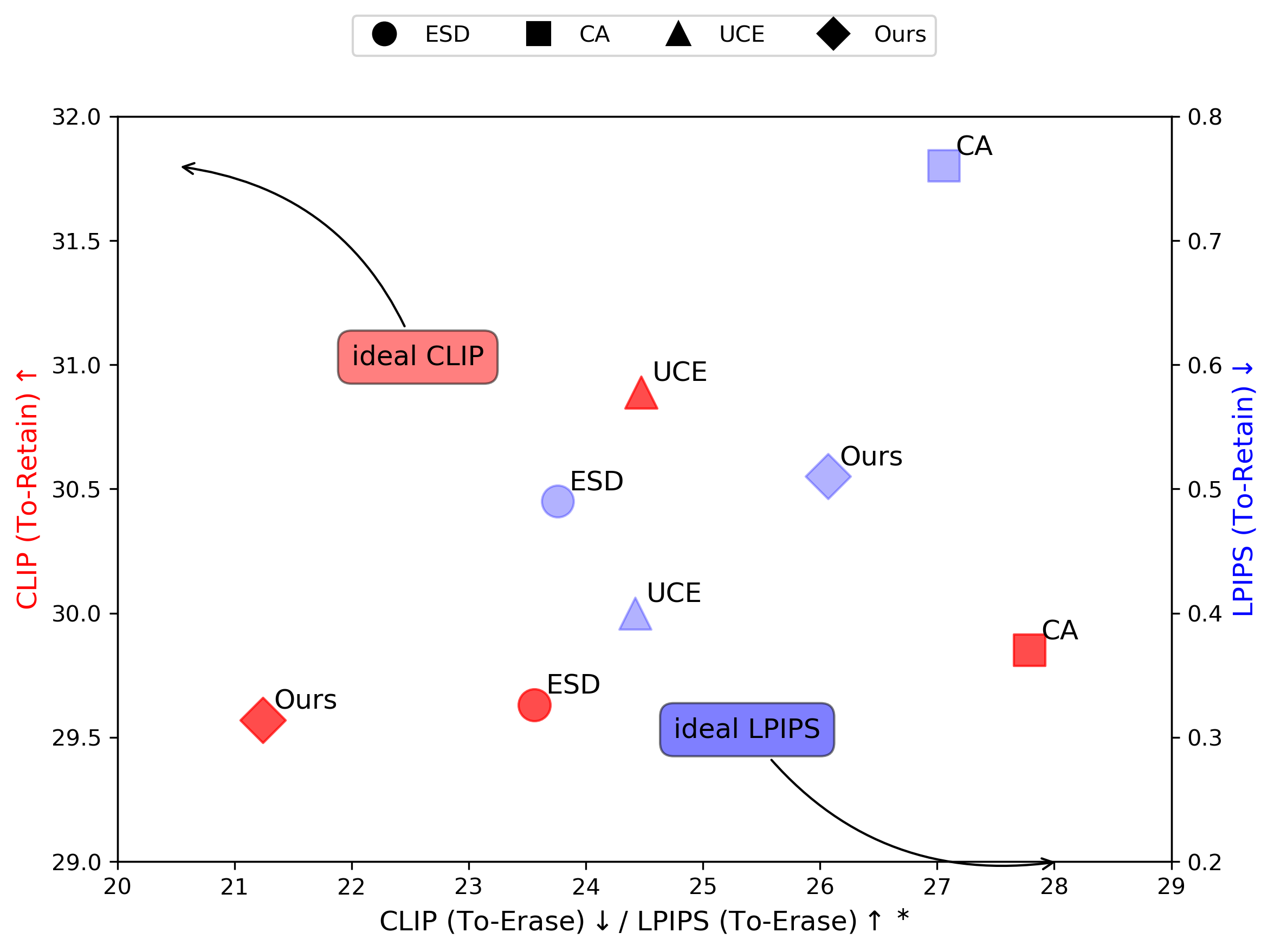}
    \caption{quantitative results of artistic style erasure.}
    \label{fig:artistic_style_erasing}
\end{figure}

It can be seen from Table \ref{tab:artistic_style_erasing} and Figure \ref{fig:artistic_style_erasing} that our method achieves the best erasing performance while maintaining a comparable preserving performance compare to the baselines. 
Specifically, our method attains the lowest CLIP score on the to-be-erased sets at 21.24, outperforming the second-best score of 23.56 achieved by ESD. Additionally, our method secures a 0.79 LPIPS score, the second-highest, following closely behind the CA method with 0.82. Concerning preservation performance, we observe that, while our method achieves a slightly higher LPIPS score than the ESD and UCE methods, suggesting some alterations compared to the original images generated by the SD model, the CLIP score of our method remains comparable to these baselines. This implies that our generated images still align well with the input prompt.

\begin{table}[h]
    \centering
    \caption{Erasing artistic style concepts.}
    \resizebox{1.0\columnwidth}{!}{%
    \begin{tabular}{llccccc}
     &  & \multicolumn{2}{c}{To Erase} &  & \multicolumn{2}{c}{To Retain}\tabularnewline
    \cline{3-4} \cline{4-4} \cline{6-7} \cline{7-7} 
     &  & CLIP $\downarrow$ & LPIPS$\uparrow$ &  & CLIP$\uparrow$ & LPIPS$\downarrow$\tabularnewline
    \hline 
    ESD &  & $23.56\pm4.73$ & $0.72\pm0.11$ &  & $29.63\pm3.57$ & $0.49\pm0.13$\tabularnewline
    CA &  & $27.79\pm4.67$ & $0.82\pm0.07$ &  & $29.85\pm3.78$ & $0.76\pm0.07$\tabularnewline
    UCE &  & $24.47\pm4.73$ & $0.74\pm0.10$ &  & $30.89\pm3.56$ & $0.40\pm0.13$\tabularnewline
    Ours &  & $21.24\pm5.56$ & $0.79\pm0.10$ &  & $29.57\pm3.72$ & $0.51\pm0.14$\tabularnewline
    \hline 
    \end{tabular}
    }
    \label{tab:artistic_style_erasing}
\end{table}

\begin{figure*}
    \centering
    \begin{subfigure}{0.32\textwidth}
        \centering
        \includegraphics[width=\columnwidth]{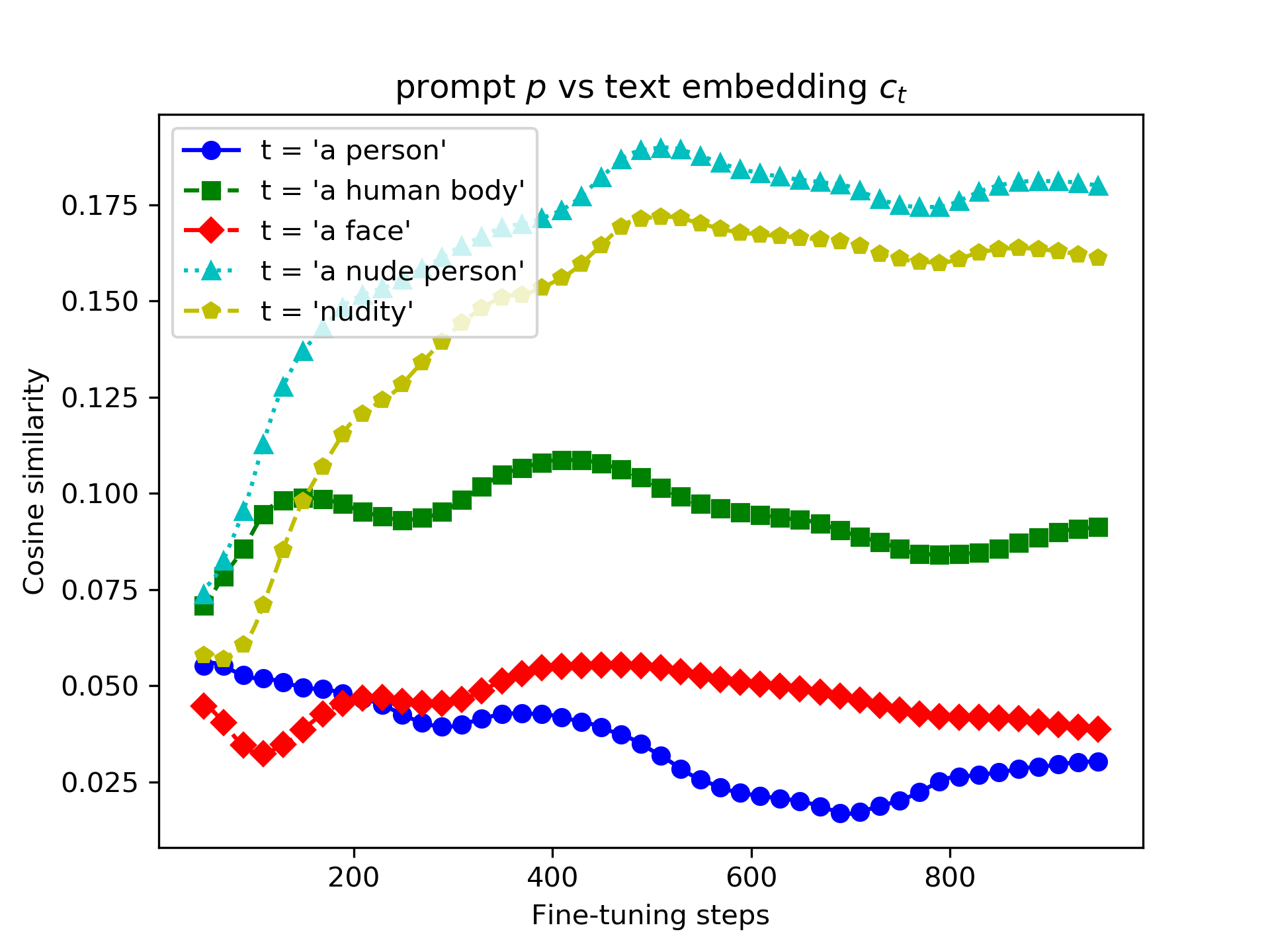}
        \caption{}
        \label{fig:cosine_vs_word}
    \end{subfigure}
    \begin{subfigure}{0.32\textwidth}
        \centering
        \includegraphics[width=\columnwidth]{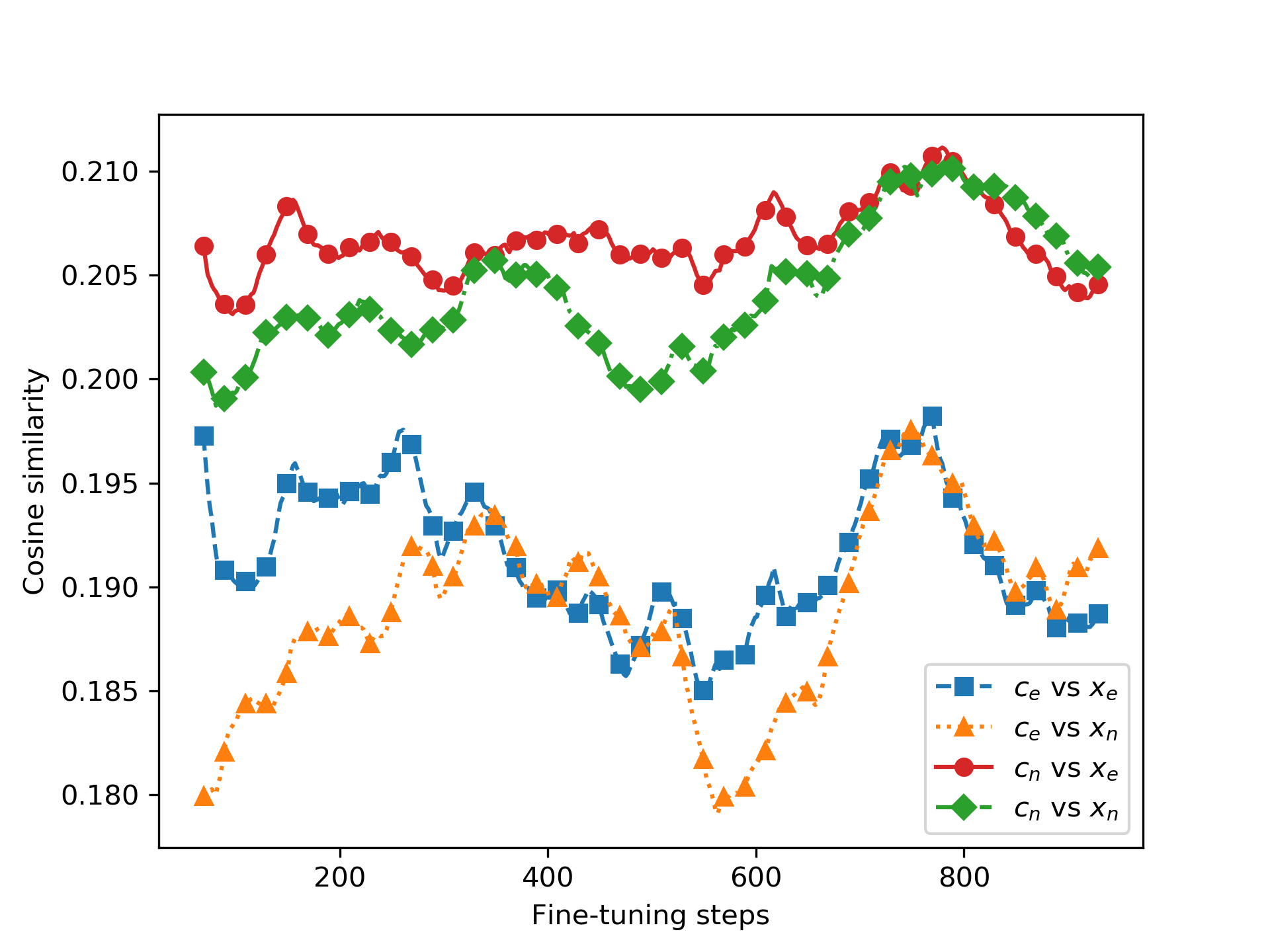}
        \caption{}
        \label{fig:cosine_vs_image}
    \end{subfigure}
    \begin{subfigure}{0.32\textwidth}
        \centering
        \includegraphics[width=\columnwidth]{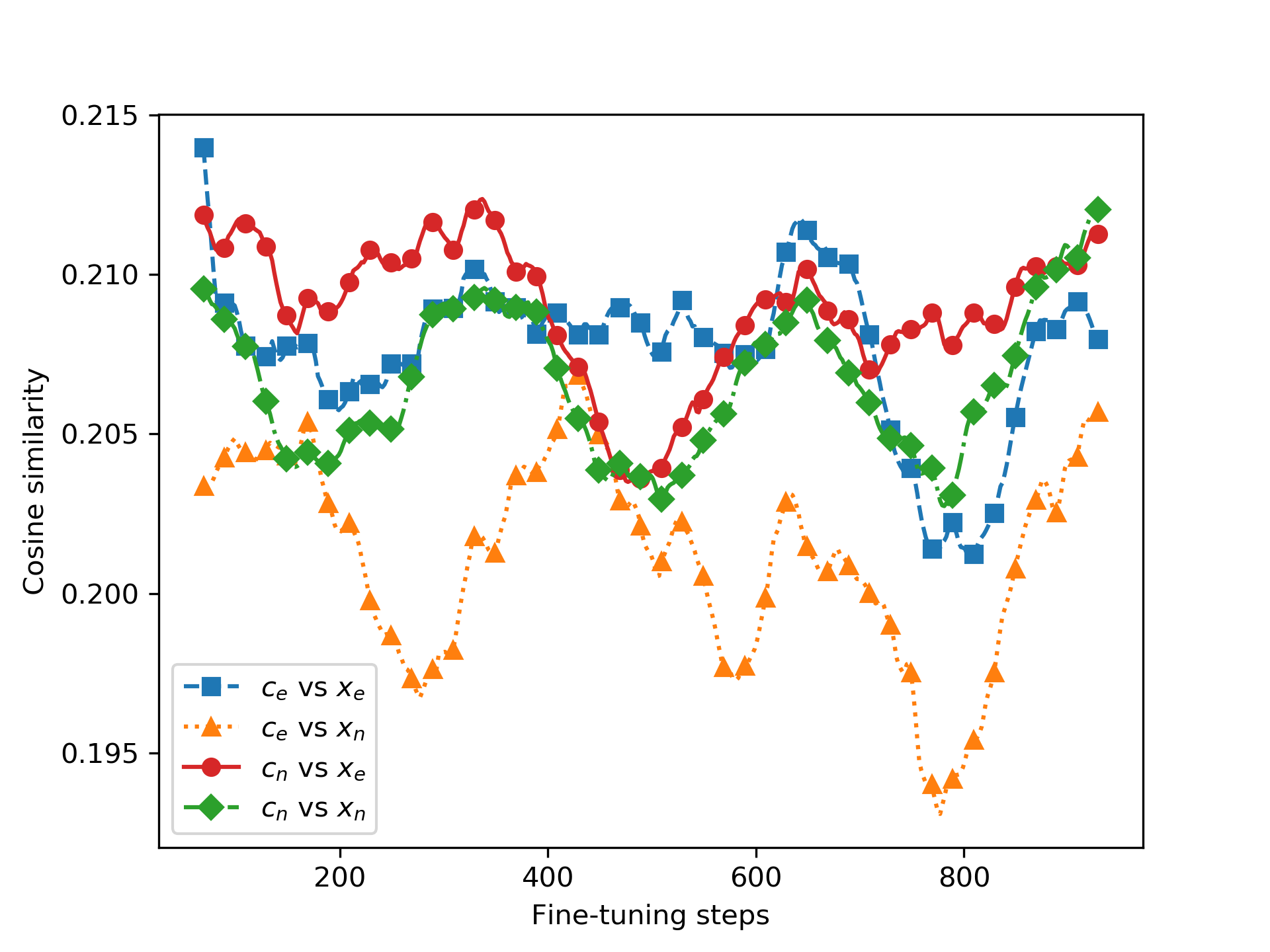}
        \caption{}
        \label{fig:cosine_vs_image_esd}
    \end{subfigure}
    \caption{Prompt's learning process (\ref{fig:cosine_vs_word}) and the cosine similarity between visual and textual features in our method (\ref{fig:cosine_vs_image})  and ESD (\ref{fig:cosine_vs_image_esd}), respectively.}
    \label{fig:understanding_prompting}
\end{figure*}

\subsection{Understanding the Prompting Mechanism}
\label{sec:understanding_prompting}

In this section, we aim to investigate the behavior of the prompting mechanism in our method, to further provide insights into the underlying mechanism of our method. 
We first analyze the learning process of the prompt, by measuring the cosine similarity between the prompt and several related textual inputs along the fine-tuning process. 
As depicted in Figure \ref{fig:cosine_vs_word}, initially, the prompt exhibits no alignment with any textual inputs.
However, through the fine-tuning process, it progressively aligns more closely with the most relevant ones, like `nudity' or `a nude person', while maintaining an uncorrelated relationship with more neutral expressions like `a person' or `a face'.
Intriguingly, although we explicitly enforce alignment with the keyword $c_e$ `nudity', the prompt aligns most with `a nude person', suggesting that it has captured the concept of nudity in a more specific sense, specifically referring to a person.

Next, we generate images $x_e$ and $x_n$ with the textual input $c_e = \text{`nudity'}$ and $c_n = \text{`a person'}$ respectively. 
We then measure the alignment between the CLIP visual and textual features of these images and their corresponding textual inputs.
As illustrated in Fig. \ref{fig:cosine_vs_image}, throughout the learning process of the prompt, there is a decline in the alignment between $c_e$ and $x_e$, indicating that the keyword $c_e$ becomes less capable of generating images with the erased concept.
Conversely, the alignment between $c_e$ and $x_n$ increases, suggesting that the keyword $c_e$ becomes more adept at generating images with neutral concepts.
Additionally, the alignment between $c_n$ and $x_n$ also increases, highlighting the preserving effect of our method.
In contrast, the alignment between pairs in the ESD method remains unstable over the learning process, as depicted in Fig. \ref{fig:cosine_vs_image_esd}, underscoring the instability of the erasure effect in ESD compared to ours.

\section{Conclusion}

In this paper, we have introduced a novel approach to concept erasure in text-to-image generative models by incorporating an additional learnable parameter prompt. This prompt helps reduce the model's dependency on generating undesirable concepts, thereby minimizing the negative impact on other unrelated concepts during the erasure process, resulting in better performance in both erasing and preserving aspects as demonstrated through extensive experiments in our paper. Furthermore, our proposed prompting mechanism exhibits high flexibility and can be extended to address other challenges involving cross-attention layers, such as continual learning. Additionally, exploring more complex prompting mechanisms, such as amortizing the prompt using a learnable function of textual embeddings, presents promising avenues for future research.

\bibliography{erasing}

\newpage

\onecolumn

\title{Hiding and Recovering Knowledge in Text-to-Image \\ Diffusion Models via Learnable Prompts\\(Supplementary Material)}
\maketitle

\section{Further Details on the Proposed Method}
\label{sec:further_details}

In this section, we will delve into discussing two different mechanisms for injecting the prompt into the cross-attention module, known as the concatenative mechanism and the additive mechanism. Their basic operations can be found in Table \ref{tab:cross_attention_mechanism}.
In text-to-image diffusion models, the cross-attention layers are positioned to integrate textual embeddings into visual generation to regulate the output's concept. 
Hence, these specific layers are as the most suitable for prompt injection, aligning with our goal.
 
\paragraph*{Concatenative Mechanism.}

In this mechanism, the prompt is concatenated with the textual embedding $\mathbf{C}$ before being used as the key and value matrix inputs to the cross-attention module.
Let $\mathbf{p} \in \mathbb{R}^{[1, m_p, d_p]}$ be the additional learnable prompt, where $m_p=k m_c$ is the prompt size and $d_p=d_c$ is the dimension of the prompt. 
The main difference compared to the original mechanism is the projected matrices $K$ and $V$ as shown in Table \ref{tab:cross_attention_mechanism}. 

Softmax normalization is applied on the last dimension of the attention score matrix $\mathbf{A}$.
In addition, the scaling factor $\sqrt{d}$ is used to prevent the attention score from being too small when the dimension of the latent vector is large \citep{vaswani2017attention}. 

One of the advantages of this mechanism is that adding the prompt does not change the mechanism of the cross-attention module, 
which means that there is no need for a new architecture and it does not interfere with the model's ability to generate good content. 
We can use the same pre-trained model and only need to modify the corresponding cross-attention module in the codebase.
With $\mathbf{p}=c$, the model can generate the same output as the original model.

Secondly, this mechanism allows us to utilize a larger prompt (theoretically, the prompt can be arbitrarily large).
By using a larger prompt, we can either remove more undesirable content or preserve more desirable content.
Experiments in Section \ref{sec:prompt_size} demonstrate that the performance of this mechanism is scalable with the size of the prompt.

\begin{figure}
    \centering
    \includegraphics[width=0.5\columnwidth]{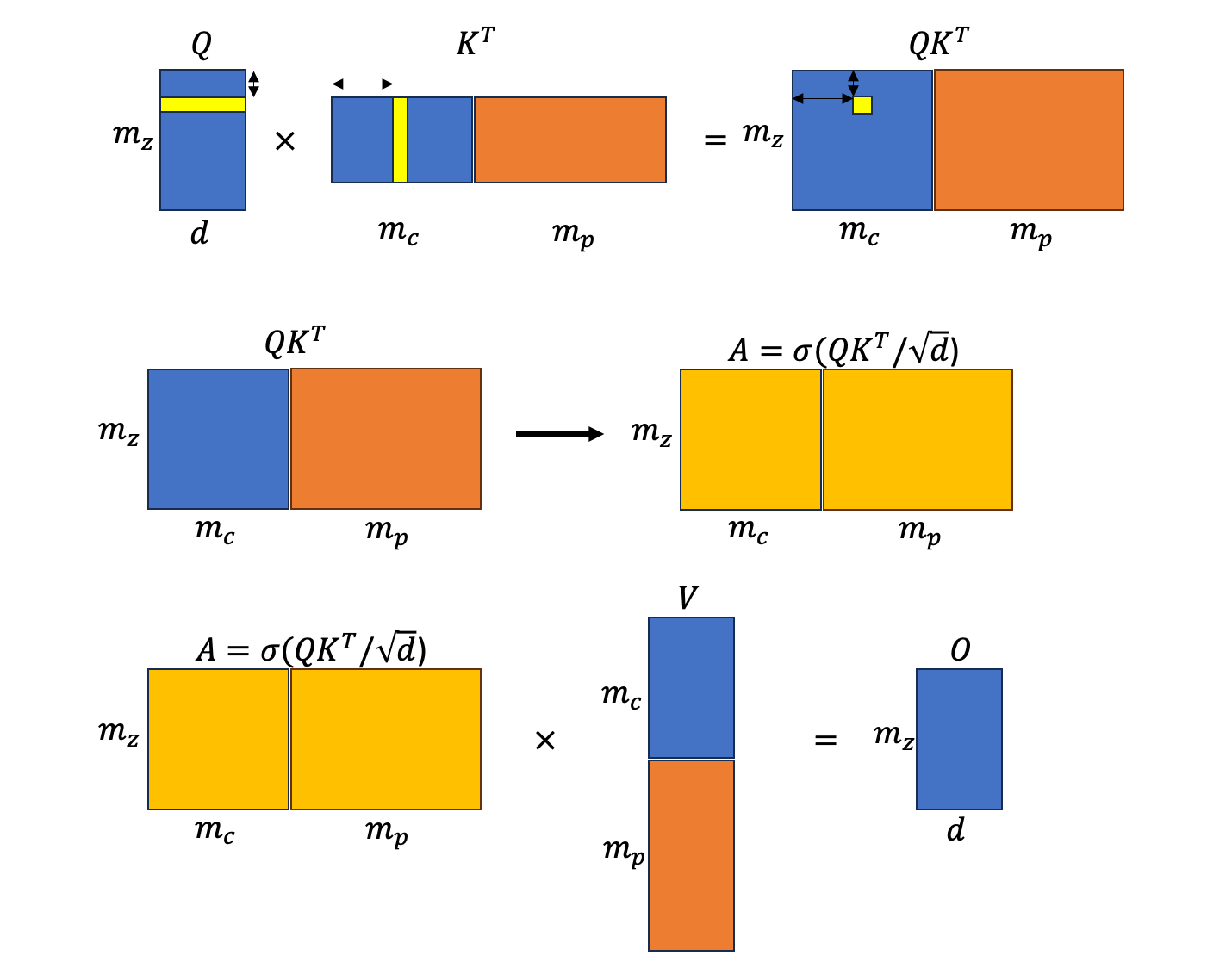}
    \caption{Illustration of Cross Attention with additional prompt.}
    \label{fig:qkv}
\end{figure}

\textit{Limitation of the concatenative mechanism:} 
The main limitation of the concatenative mechanism is that it relies on softmax normalization to distribute the attribution from the additional prompt to the entire textual path.
This issue is also illustrated in Figure \ref{fig:qkv}. 
Because of the linearity of the projection operation, the matrices $\mathbf{K}$ and $\mathbf{V}$ can be decomposed into two parts separately, 
the projections of original textual embedding $\mathbf{C}$ (blue color) and prompt $\mathbf{P}$ (orange color), i.e., $\mathbf{K} = [\mathbf{W}_k \mathbf{C}, \mathbf{W}_k \mathbf{P}], \mathbf{V} = [\mathbf{W}_v \mathbf{C}, \mathbf{W}_v \mathbf{P}]$. 
As a result, the dot-product between the query $\mathbf{Q}$ and the key $\mathbf{K}$ can also be decomposed into two disjointed parts, i.e., $\mathbf{Q}\mathbf{K}^T = [\mathbf{Q} (\mathbf{W}_k \mathbf{C})^T, \mathbf{Q} (\mathbf{W}_k \mathbf{P})^T]$.
Without the softmax normalization, the output of the cross-attention module will be just the sum of the two disjointed parts, 
which is not desirable. 
The softmax normalization applied on the last dimension of the dot-product score matrix $\mathbf{Q}\mathbf{K}^T$ helps to distribute the attribution from the prompt to the entire textual path, 
enabling the model to attend to the additional prompt better.
However, as the prompt size increases, the softmax normalization will distribute the attribution from the prompt to the entire textual path more evenly, 
which can lead to the model's inability to attend to the prompt effectively, as shown in Section \ref{sec:prompt_size}.

\paragraph*{Additive Mechanism.}
This mechanism injects the additional prompt by directly adding it to the textual embedding $\mathbf{C}$ before being used as the key and value matrix inputs to the cross-attention module.
This retains the same advantages as the concatenative mechanism, i.e., it does not change the mechanism of the cross-attention module. 
It also permits a deeper integration of the prompt into the textual path, which allows the model to attend to the prompt more effectively.
However, a limitation of this mechanism is that it is not scalable since its size is fixed to the size of the textual embedding.
We compare the performance of two mechanisms in Section \ref{sec:prompting_mechanism}.

\paragraph*{Alternative Prompting Mechanism.}

Beyond the two aforementioned mechanisms, we acknowledge that there are several potential prompting mechanisms that can be used to modify the cross-attention module.
For example, we can inject the prompt before the text encoder, by using a learnable word embedding vector associated with a special token $S^*$ to represent the prompt as in Textual Inversion \citep{gal2022image}.
We can also amortize the prompt by using a learnable function to generate the prompt from the textual embedding, i.e., $\mathbf{p} = f(\mathbf{c})$. 
We leave the exploration of these mechanisms for future work.

\begin{table}[!]
    \centering
    \caption{Cross-Attention Mechanisms}
    \resizebox{1.0\columnwidth}{!}{%
    \addtolength{\tabcolsep}{1pt} 
    \begin{tabular}{llllllllll}
    \hline 
     &  & Original &  &  & Concatenative &  &  & Addititive & \tabularnewline
    \cline{3-4} \cline{4-4} \cline{6-7} \cline{7-7} \cline{9-10} \cline{10-10} 
     &  & Operation & Dim &  & Operation & Dim &  & Operation & Dim\tabularnewline
    \hline 
    $Q$ &  & $W_{q}Z$ & $b\times m_{z}\times d$ &  & $W_{q}Z$ & $b\times m_{z}\times d$ &  & $W_{q}Z$ & $b\times m_{z}\times d$\tabularnewline
    $K$ &  & $W_{k}C$ & $b\times m_{c}\times d$ &  & $W_{k}\;\text{cat}(C,\text{repeat}(p,b))$ & $b\times(m_{c}+m_{p})\times d$ &  & $W_{k}\;(C+\text{repeat}(p,b))$ & $b\times m_{c}\times d$\tabularnewline
    $V$ &  & $W_{v}C$ & $b\times m_{c}\times d$ &  & $W_{v}\;\text{cat}(C,\text{repeat}(p,b))$ & $b\times(m_{c}+m_{p})\times d$ &  & $W_{v}\;(C+\text{repeat}(p,b))$ & $b\times m_{c}\times d$\tabularnewline
    $A$ &  & $\sigma(QK^{T}/\sqrt{d})$ & $b\times m_{z}\times m_{c}$ &  & $\sigma(QK^{T}/\sqrt{d})$ & $b\times m_{z}\times(m_{c}+m_{p})$ &  & $\sigma(QK^{T}/\sqrt{d})$ & $b\times m_{z}\times m_{c}$\tabularnewline
    $O$ &  & $AV$ & $b\times m_{z}\times d$ &  & $AV$ & $b\times m_{z}\times d$ &  & $AV$ & $b\times m_{z}\times d$\tabularnewline
    \hline 
    \end{tabular}
    \addtolength{\tabcolsep}{-1pt} 
    }
    \label{tab:cross_attention_mechanism}
\end{table}

\section{Further Experiments}
\label{sec:further_experiments}

\subsection{Ablation Study}
\label{sec:ablation_study}

\subsubsection{Concatenative vs Additive Mechanism}
\label{sec:prompting_mechanism}

In this experiment, we conducted a comparison of erasing performance between two mechanisms. 
The evaluation was performed on a subset of 5 classes from the Imagenette dataset, including `Cassette Player', `Church', `Garbage Truck', `Parachute', and `French Horn'. 
Additionally, we assessed the erasing performance in a nudity concept setting using the I2P dataset with the NER at a threshold of 0.5 as the erasure metric.

The results, presented in Table \ref{tab:prompting_mechanism_size}, indicate that the concatenative prompting mechanism outperforms the additive prompting mechanism in terms of erasing performance. 
This is evident in the 2.44\% increase in ESR-1 and 3.16\% increase in ESR-5, as well as a 0.3\% decrease in NER. 
However, it is worth noting that the concatenative prompting mechanism is less effective in preserving unrelated concepts, 
as indicated by a drop of 2.8\% in PSR-1 compared to the additive prompting mechanism.

While additive prompting theoretically provides a deeper integration between the prompt and the real textual input, 
the concatenative prompting mechanism has demonstrated greater effectiveness in erasing the target concept. 
Furthermore, its scalability allows for varying prompt sizes, a feature discussed in the next section. 
As a result of its superior erasing performance, we adopt the concatenative prompting mechanism as the default setting in all other experiments.

\begin{table}[ht]
    \centering
    \caption{Analytical results to different prompting mechanisms and prompt size.}
    \resizebox{0.5\columnwidth}{!}{%
    \begin{tabular}{cccccc}
        Method & ESR-1$\uparrow$ & ESR-5$\uparrow$ & PSR-1$\uparrow$ & PSR-5$\uparrow$ & NER$\downarrow$\tabularnewline
        \hline 
        Additive & 96.40 & 92.32 & 84.48 & 97.92 & 1.7\tabularnewline
        Concat & 98.84 & 95.48 & 81.68 & 97.56 & 2.0\tabularnewline
        \hline 
        k=1 & 98.60 & 96.04 & 84.76 & 97.56 & 2.17\tabularnewline
        k=10 & 98.84 & 95.48 & 81.68 & 97.56 & 1.70\tabularnewline
        k=100 & 99.68 & 97.08 & 82.68 & 96.84 & \textbf{1.15}\tabularnewline
        k=200 & 99.60 & 96.80 & 77.24 & 94.16 & 1.49\tabularnewline        
        \hline 
    \end{tabular}
    }
    \label{tab:prompting_mechanism_size}
\end{table}

\subsubsection{Effect of Prompt Size}
\label{sec:prompt_size}

In this experiment, we explore the influence of prompt size on erasing performance by systematically varying the parameter $k$ within the range of 1 to 200.
The experimental setup mirros the previous experiment, where we focus on erasing object-related concepts with 5 classes from the Imagenette dataset, 
and also erasing nudity concepts with the I2P dataset.

It can be seen from the results in Table \ref{tab:prompting_mechanism_size} that the erasing performance increases as the prompt size becomes larger, 
but becomes saturated once the prompt size becomes sufficiently large. 
Specifically, ESR-1 improves from 98.60\% to 99.68\%, and ESR-5 from 96.04\% to 97.08\% as the prompt size increases from 1 to 100. 
Similarly, the NER score decreases from 2.2 to 1.1 within the same range, indicating a consistent impact of prompt size across different types of concepts.
However, the erasing performance is accompanied by a trade-off in preserving performance, 
as PSR-1 decreases from 84.76\% to 77.24\% when the prompt size increases from 1 to 200.
The observed saturation in erasing performance for larger prompt sizes can be attributed to the softmax function in the cross-attention module, 
which becomes increasingly uniform and small as the prompt size grows. 
This makes it more challenging for the model to distinctly focus on the specific concept for erasure.

\subsubsection{Influence of Hyper-parameter}
\label{sec:hyperparam_lambda}

In this experiment, we investigate the influence of the hyper-parameter $\lambda$ on erasing performance of our method. 
We conduct the experiment to erase object-related concepts, using the same experimental setup as described in Section 4.6 of the main paper. 
Specifically, we focus on a subset of 5 classes from the Imagenette dataset, including `Cassette Player', `Church', `Garbage Truck', `Parachute', and `French Horn' as the concepts to be erased, 
while preserving the remaining classes. We vary the hyper-parameter $\lambda$ from 0.01 to 10.0.
It is worth noting that the hyper-parameter $\lambda$ is utilized to introduce the prompt in the knowledge removal stage, as described in Section 3.2 of the main paper. 
Therefore, it must be strictly positive, i.e., $\lambda > 0$.

\begin{figure}
    \centering
    \includegraphics[width=0.5\columnwidth]{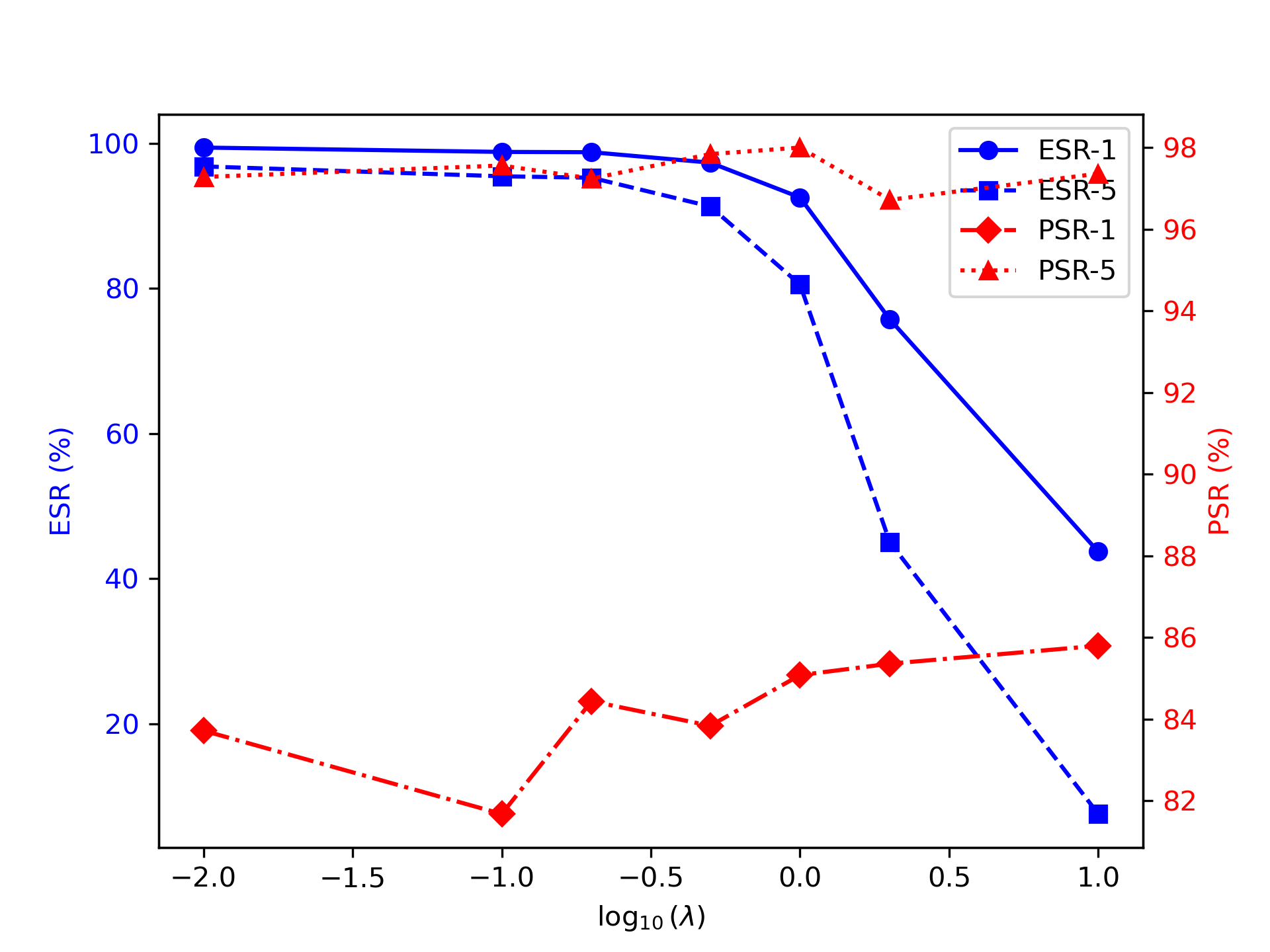}
    \caption{Impact of the hyper-parameter $\lambda$ on the erasing performance.}
    \label{fig:hyperparam_lambda}
\end{figure}

The results depicted in Figure \ref{fig:hyperparam_lambda} reveal a clear decreasing trend in erasing performance as the hyper-parameter $\lambda$ increases, while the preserving performance exhibits a slight increase.
Specifically, the erasing performance peaks at $\lambda = 0.01$, with an ESR-5 of 96.8\%, and drops significantly to 80.6\% at $\lambda = 1.0$, and to around 40\% at $\lambda = 10.0$.
Conversely, the preserving performance, measured by PSR-5, increases from 97.3\% to 98.0\% as $\lambda$ increases.
The PSR-1 also exhibits a similar trend, increasing from 81.6\% to around 86\% as $\lambda$ varies from 0.01 to 10.0.

This result aligns with our analysis in Section 3.2 of the main paper, where the hyper-parameter $\lambda$ is employed to control the trade-off between erasing and preserving performance.
In the knowledge removal stage, the $L_2$ term serves as a regularization term to minimize the change in the model's parameters, 
thereby preserving the knowledge encoded in the prompt of the erased concepts learned from the knowledge transfer stage. 
Therefore, a larger $\lambda$ encourages the model to preserve the knowledge in the prompt more strongly, leading to smaller changes in the model's parameters and better preserving performance, but worse erasing performance. 
In other experiments, we use $\lambda = 0.1$ as the default value for the hyper-parameter $\lambda$. 

\subsubsection{Where to Inject Prompt}
\label{sec:prompt_layer}

In this paper, we have introduced two different prompting mechanisms: concatenative and additive prompting. 
While in the additive prompting, we add the prompt $\mathbb{p}$ directly to the textual embedding $c$ as in Table \ref{tab:cross_attention_mechanism}, 
which can be understood as injecting the prompt in entire cross-attention layers of the U-Net, 
in the concatenative prompting, we need to specify where to inject the prompt. 

In this experiment, we explore the influence of prompt injection at different layers of the model on erasing performance within the U-Net architecture of the Stable Diffusion model. 
The U-Net comprises three main components: the down-sample blocks, the middle block, and the up-sample blocks. 
Each of these components includes multiple cross-attention layers that can be used to inject the textual/conditional input, 
except for the middle block, which contains only one cross-attention layer.

\begin{table}[]
    \centering
    \caption{Where to inject the prompt.}
    \resizebox{0.5\columnwidth}{!}{%
    \begin{tabular}{ccccc}
        layer & ESR-1$\uparrow$ & ESR-5$\uparrow$ & PSR-1$\uparrow$ & PSR-5$\uparrow$ \tabularnewline
        \hline 
        mid & 98.84 & 95.48 & 81.68 & 97.56 \tabularnewline
        mid-up & 99.64 & 97.92 & 75.68 & 95.12 \tabularnewline
        down-mid-up & 99.65 & 98.36 & 59.04 & 86.28 \tabularnewline
        \hline 
    \end{tabular}
    }
    \label{tab:prompt_layer}
\end{table}

We compare three different settings: injecting the prompt at the middle block, the middle and up-sample blocks, and the down-middle-up sample blocks (i.e., all cross-attention layers in the U-Net), 
with the same experimental setup as in previous experiments.
The results in Table \ref{tab:prompt_layer} indicate that injecting the prompt at all cross-attention layers in the U-Net yields the best erasing performance, 
albeit with a significant drop in preserving performance.

It is noteworthy that as the number of cross-attention layers used for prompt injection increases, erasing performance improves at the expense of preserving performance. 
The optimal trade-off between erasing and preserving performance is achieved by injecting the prompt at the middle block only. 
This setting was consequently chosen as the default for all subsequent experiments. 
It strikes a balance, demonstrating effective erasure while still preserving relevant elements in the input.

\subsubsection{Further Results on Erasing Artistic Style Concepts}

\paragraph{How to systematically evaluate the erasure performance?}

In this paper, we have conducted three sets of experiments to assess the performance of our proposed method against other erasure baselines. 
In the initial two sets, targeting the erasure of object-related and nudity concepts, we employed pre-trained detectors like ResNet-50 \citep{resnet} and Nudenet \citep{nudenet2019} to identify the presence of these concepts in the generated images. 
This systematic approach enabled us to evaluate the erasure performance rigorously. 
However, in the final experiment set, aimed at erasing artistic style concepts, we encountered a challenge: the absence of a pre-trained detector capable of accurately assessing the presence of an artistic style in the generated images.
To address this challenge, previous studies \citep{gandikota2023erasing} proposed human evaluations, which are subjective and time-consuming. 

\begin{table}[]
    \centering
    \caption{CLIP alignment score measured on the original SD model.}
    \resizebox{0.75\columnwidth}{!}{%
    \begin{tabular}{lccccccc}
        & Content \& Artist &  &  & Artist &  &  & Content\tabularnewline
       \hline 
       $\text{Kelly McKernan}$ & $31.47\pm2.58$ &  &  & $27.67\pm2.73$ &  &  & $29.69\pm2.43$\tabularnewline
       $\text{Tyler Edlin}$ & $30.63\pm2.22$ &  &  & $23.67\pm1.24$ &  &  & $30.12\pm2.49$\tabularnewline
       $\text{Kilian Eng}$ & $29.87\pm2.64$ &  &  & $25.08\pm1.31$ &  &  & $30.54\pm2.36$\tabularnewline
       $\text{Thomas Kinkade}^{\star}$ & $34.63\pm1.96$ &  &  & $31.13\pm2.38$ &  &  & $31.09\pm2.22$\tabularnewline
       $\text{Ajin: Demi Human}^{\star}$ & $30.70\pm2.55$ &  &  & $27.65\pm3.24$ &  &  & $25.38\pm2.77$\tabularnewline
       $\text{VanGogh}^{\star}$ & $33.66\pm2.41$ &  &  & $30.36\pm1.17$ &  &  & $28.62\pm3.28$\tabularnewline
       \hline 
       \end{tabular}
    }
    \label{tab:artist_sd}
\end{table}

In this experiment, we explored the use of the CLIP alignment score as an alternative metric to evaluate erasure performance. 
Initially, we generated 1200 images from the original SD model using lengthy, specifically designed prompts (credited to \citep{gandikota2023erasing}) to capture images with the artistic style of a particular artist. 
Subsequently, we measured the CLIP alignment score between the generated images and three different textual inputs: the full prompt containing both the content and the artist name, the artist name alone, and the content alone. 
The results presented in Table \ref{tab:artist_sd} revealed intriguing insights. 
On one hand, when measuring based solely on the artist name, the CLIP alignment score was consistently the lowest in all cases, except for Thomas Kinkade, Ajin: Demi Human, and VanGogh. 
Conversely, when measuring based solely on the content, the CLIP alignment score was relatively higher. 
Lastly, when considering the full prompt, inclusive of both the content and the artist name, the CLIP alignment score was consistently the highest in all cases, except for Kilian Eng. 
This suggests that, from the CLIP's perspective, the generated images may not align well with just the artist name, but they exhibit strong alignment with the full prompt, encompassing both the content and the artist name. 
Consequently, to evaluate erasure performance, we can leverage CLIP as a zero-shot classifier, as highlighted in Section \ref{sec:artistic_style_concepts}.

\paragraph{Qualitative Results.}

\begin{figure}
    \centering
    \includegraphics[width=\columnwidth]{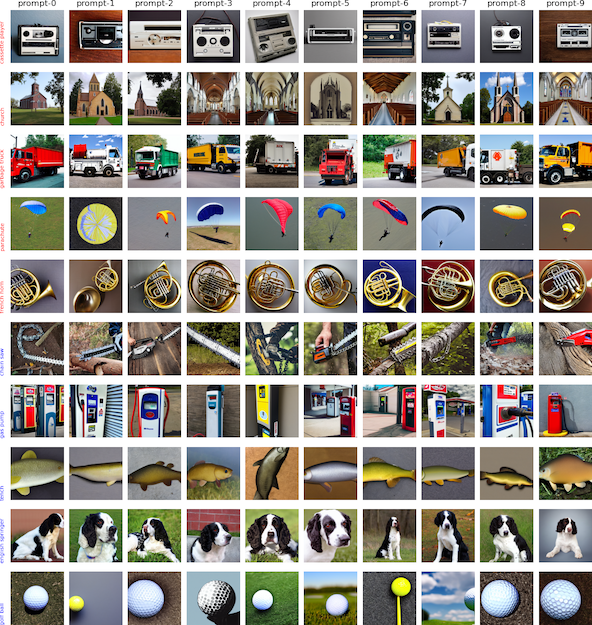}

    \caption{Generated images from the original model. 
    Five first rows are to-be-erased objects (marked by red text) and the rest are to-be-preserved objects. 
    Each column represents different random seeds.  
    }
    \label{fig:object_erasure_sd}
\end{figure}

\begin{figure}
    \centering
    \includegraphics[width=\columnwidth]{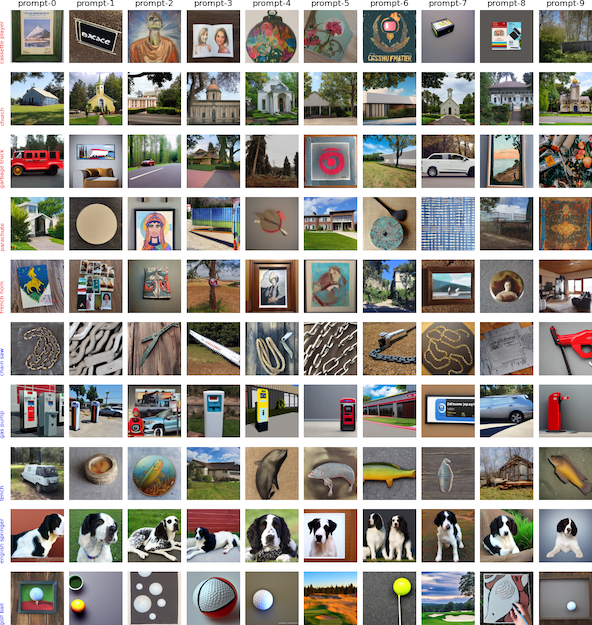}

    \caption{Erasing objects using ESD. 
    Five first rows are to-be-erased objects (marked by red text) and the rest are to-be-preserved objects. 
    Each column represents different random seeds.  
    }
    \label{fig:object_erasure_esd}
\end{figure}

\begin{figure}
    \centering
    \includegraphics[width=\columnwidth]{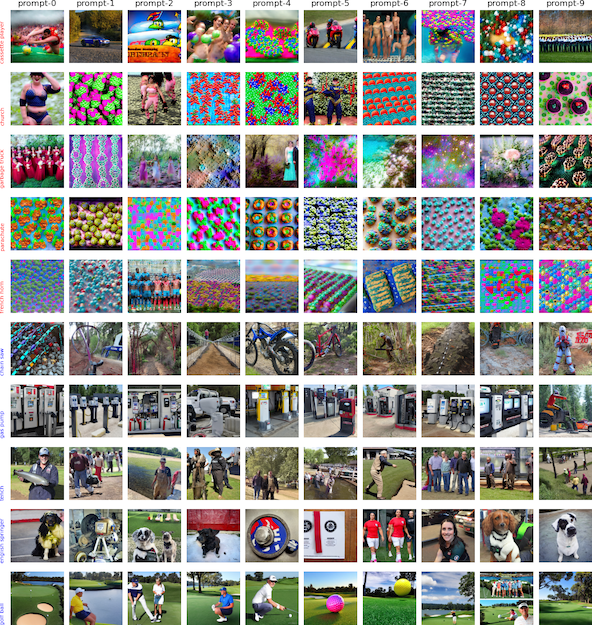}

    \caption{Erasing objects using UCE. 
    Five first rows are to-be-erased objects (marked by red text) and the rest are to-be-preserved objects. 
    Each column represents different random seeds.  
    }
    \label{fig:object_erasure_uce}
\end{figure}

\begin{figure}
    \centering
    \includegraphics[width=\columnwidth]{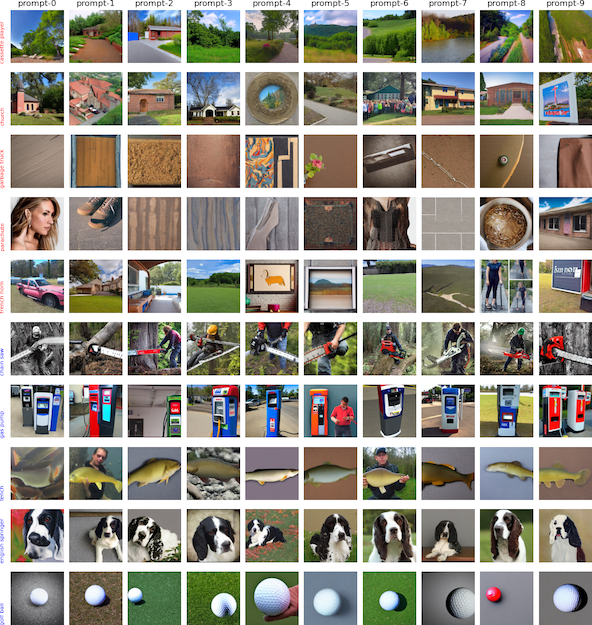}

    \caption{Erasing objects using our method (KPOP). 
    Five first rows are to-be-erased objects (marked by red text) and the rest are to-be-preserved objects. 
    Each column represents different random seeds.  
    }
    \label{fig:object_erasure_ours}
\end{figure}

\begin{figure}
    \centering
    \begin{subfigure}{0.49\columnwidth}
        \centering
        \includegraphics[width=\columnwidth]{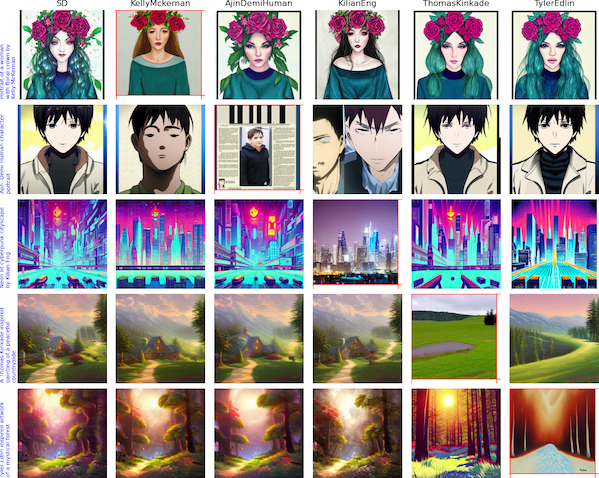}
        \caption{KPOP}
        \label{fig:ours-1}
    \end{subfigure}
    \begin{subfigure}{0.49\columnwidth}
        \centering
        \includegraphics[width=\columnwidth]{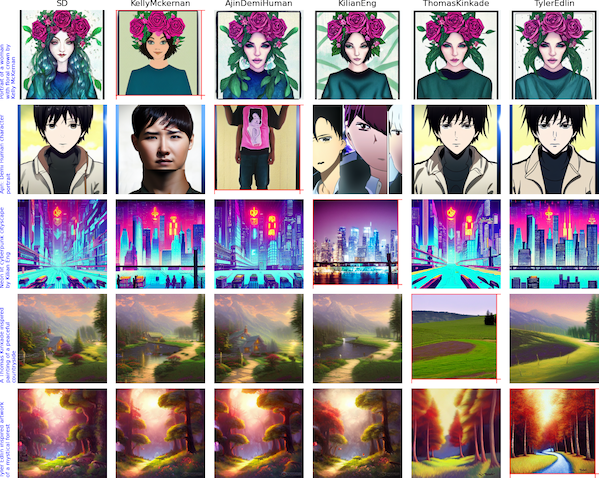}
        \caption{ESD}
        \label{fig:esd-1}
    \end{subfigure}

    \begin{subfigure}{0.49\columnwidth}
        \centering
        \includegraphics[width=\columnwidth]{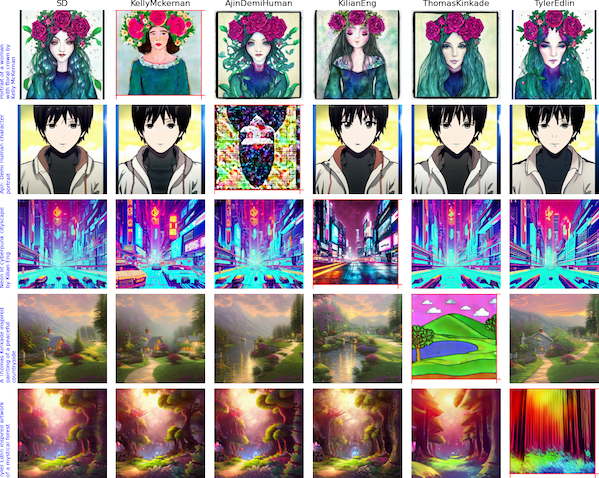}
        \caption{UCE}
        \label{fig:uce-1}
    \end{subfigure}
    \begin{subfigure}{0.49\columnwidth}
        \centering
        \includegraphics[width=\columnwidth]{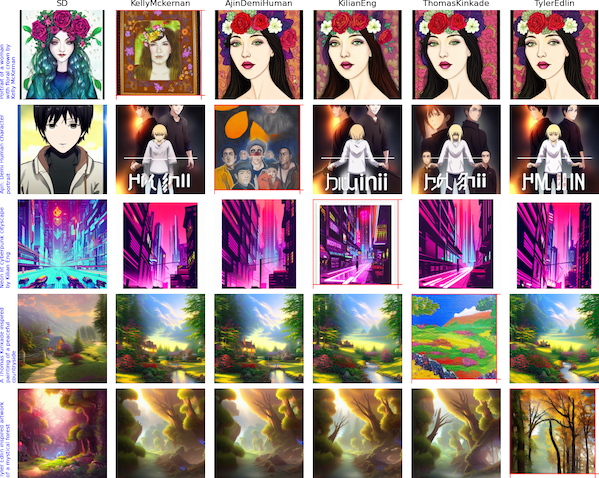}
        \caption{CA}
        \label{fig:ca-1}
    \end{subfigure}

    \caption{Erasing artistic style concepts. Each column represents the erasure of a specific artist, except the first column which represents the generated images from the original SD model. 
    Each row represents the generated images from the same prompt but with different artists. 
    The ideal erasure should result in the change in the diagonal pictures (marked by a red box) compared to the first column, while the off-diagonal pictures should remain the same. 
}
    \label{fig:artist_erasure_1}
\vspace*{-5mm}
\end{figure}

\begin{figure}
    \centering
    \begin{subfigure}{0.49\columnwidth}
        \centering
        \includegraphics[width=\columnwidth]{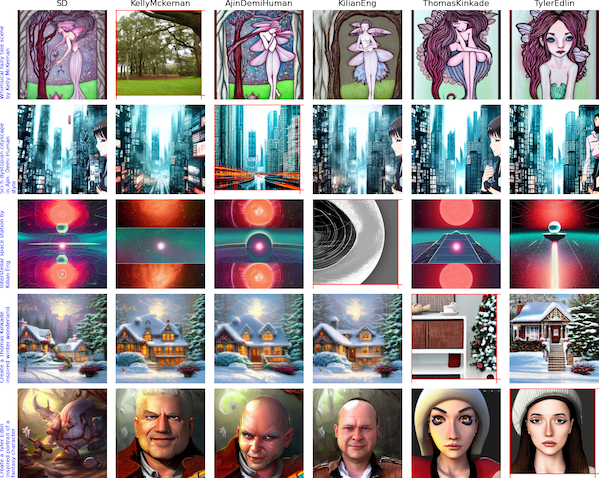}
        \caption{KPOP}
        \label{fig:ours-2}
    \end{subfigure}
    \begin{subfigure}{0.49\columnwidth}
        \centering
        \includegraphics[width=\columnwidth]{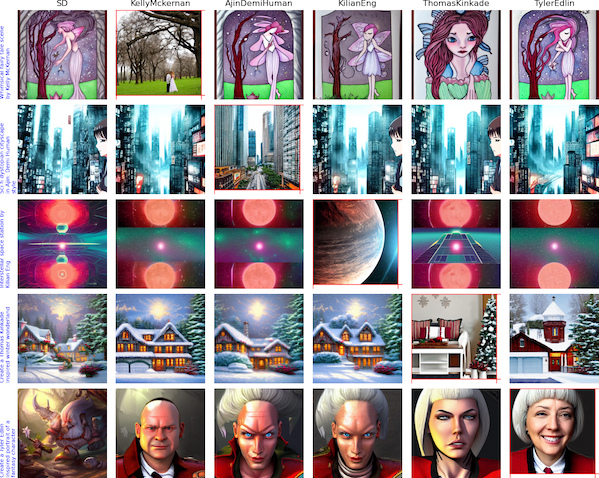}
        \caption{ESD}
        \label{fig:esd-2}
    \end{subfigure}

    \begin{subfigure}{0.49\columnwidth}
        \centering
        \includegraphics[width=\columnwidth]{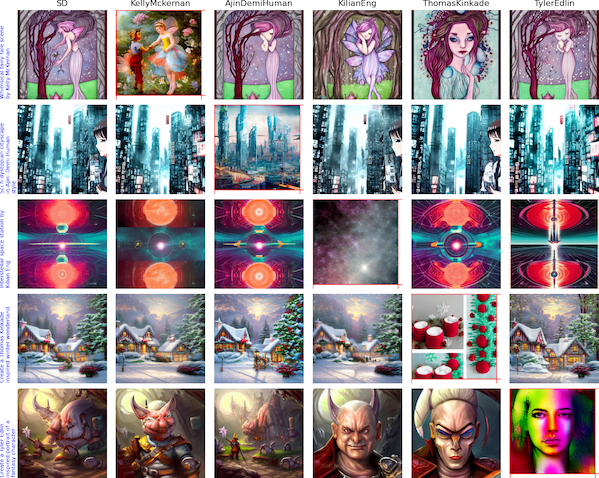}
        \caption{UCE}
        \label{fig:uce-2}
    \end{subfigure}
    \begin{subfigure}{0.49\columnwidth}
        \centering
        \includegraphics[width=\columnwidth]{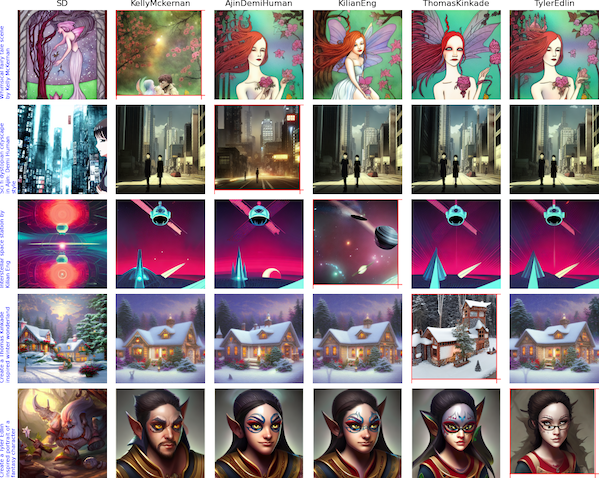}
        \caption{CA}
        \label{fig:ca-2}
    \end{subfigure}

    \caption{Erasing artistic style concepts (continue). Each column represents the erasure of a specific artist, except the first column which represents the generated images from the original SD model. 
    Each row represents the generated images from the same prompt but with different artists. 
    The ideal erasure should result in a change in the diagonal pictures (marked by a red box) compared to the first column, while the off-diagonal pictures should remain the same.
    }
    \label{fig:artist_erasure_2}
\vspace*{-5mm}
\end{figure}

In addition to the quantitative results reported in Section \ref{sec:artistic_style_concepts}, 
we provide further qualitative results as shown in series of figures from Figure \ref{fig:object_erasure_sd} to Figure \ref{fig:artist_erasure_2} to illustrate the erasure performance of our method and the baselines. 
Because of our internal policy on publishing sensitive content like nudity, we are able to provide results for the erasure of artistic style concepts and object-related concepts only.

\end{document}